\documentclass[review]{elsarticle}
\usepackage[subrefformat=parens,labelformat=parens]{subfig}
\usepackage{amsmath}
\usepackage{soul}
\usepackage{amssymb}
\usepackage{tabularx}
\journal{Journal of Pattern Recognition}
\usepackage{times}  
\usepackage{helvet} 
\usepackage{courier}  
\usepackage[]{url}
\usepackage{epsfig}
\usepackage{graphicx}
\usepackage{cite}
\usepackage{multirow}
\usepackage{multicol}
\usepackage{amsmath}
\usepackage{amssymb}
\usepackage{latexsym}
\usepackage{tabularx}
\usepackage{booktabs}
\usepackage{color}
\usepackage{makecell}
\usepackage{booktabs}
\usepackage{array}
\usepackage{bm}
\usepackage{diagbox}
\usepackage{url}
\usepackage{hyperref}

\begin{document}
\begin{frontmatter}
\title{Meta Transfer Learning for Emotion Recognition}


\author[1]{Dung Nguyen}
\author[1]{Sridha Sridharan} 

\author[2]{Duc Thanh Nguyen}

\author[1]{Simon Denman}

\author[1]{David Dean} 
\author[1]{Clinton Fookes}  


\address[1]{Speech, Audio, Image and Video Technology (SAIVT) Laboratory - Queensland University of Technology}
\address[2]{School of Information Technology - Deakin University}
\begin{abstract}
Deep learning has been widely adopted in automatic emotion recognition and has lead to significant progress in the field. However, due to insufficient annotated emotion datasets, pre-trained models are limited in their generalization capability and thus lead to poor performance on novel test sets. To mitigate this challenge, transfer learning performing fine-tuning on pre-trained models has been applied. However, the fine-tuned knowledge may overwrite and/or discard
important knowledge learned from pre-trained models. In this paper, we address this issue by proposing a PathNet-based transfer learning method that is able to transfer emotional knowledge learned from one visual/audio emotion domain to another visual/audio emotion domain, and transfer the emotional knowledge learned from multiple audio emotion domains into one another to improve overall emotion recognition accuracy. To show the robustness of our proposed
system, various sets of experiments for facial expression recognition and speech emotion recognition task on three emotion datasets: SAVEE, EMODB, and eNTERFACE have been carried out. The experimental results indicate that our proposed system is capable of improving the performance of emotion recognition, making its performance substantially superior to the recent proposed fine-tuning/pre-trained models based transfer learning methods.
\end{abstract}

\end{frontmatter}
\section{Introduction}
\label{sec1}

Emotions of humans manifest in their facial expressions, voice, gestures, and posture. An accurate emotion recognition system based on one or a combination of these modalities would be useful in various applications including surveillance, medical, robotics, human computer interaction, affective computing, and automobile safety \citep{7926723}. Researchers in this area have focused mainly
in the area of facial expression recognition to build reliable emotion recognition
systems. This is still a challenging problem since very subtle emotional changes manifested in the facial expression could go undetected \citep{7926723}. Recently several approaches based on deep learning techniques have contributed to progressing in this area \citep{Fan:2016:VER:2993148.2997632,Abbasnejad_2017_ICCV,8015016}.

In addition to facial expression stream, speech signals, which are regarded as one of the most natural media of human communication, carry both the
contents of explicit linguistic and the information of implicit paralinguistic expressed by a speaker \citep{8085174}. Due to this rich information contained
in the speech, over the last two decades numerous studies and efforts have been devoted to progressing approaches, focusing on automatic and accurate detection of human emotions from speech signals \citep{8085174}. Speech emotion recognition is presently playing an essential role in a wide range of
applications such as automobile safety, surveillance, human computer interaction, and robotics, and is attracting a great deal of attention within the affective computing research community \citep{NGUYEN2018}.

To develop solutions for speech emotion recognition, a number of methodologies have been proposed, in which researchers have primarily applied the
use of hand engineered features based on the acoustic and paralinguistic information \citep{6023178}. Nevertheless, such hand-designed features seem not to be discriminative enough to boost the performance of speech emotion recognition \citep{8085174}. Recently,
algorithms based on deep learning techniques, which are capable of automatically learning features and also capable of modelling high-level information, have been the focus of most recent research and are gaining prominence. 

Although the aforementioned deep learning approaches have made a great contribution to progressing the emotion recognition area, we pointed out a key issue that plagues the advancement of emotion recognition research; e.g., the lack of sufficient quantities of annotated emotion data. This issue has become more critical with the advent of deep learning techniques which promise major improvements in emotion recognition accuracy in both single and multi-modal settings; yet we are unable to exploit the full potential of deep learning for emotion recognition due to the scarcity of annotated emotion data; deep learning techniques require large amounts of data for training. Transfer learning method, which commonly fine-tunes pre-trained/off-the-shelf CNN models on emotion dataset, have been widely investigated to overcome this problem. However, the representational features that are unrelated to emotion are still retained in off-the-shelf/pre-trained models and the extracted features are also vulnerable to identity variations in these approaches, leading to degrading the performance of the emotion recognition system fine-tuning off-the-shelf/pre-trained models on the emotion dataset.

To resolve these drawbacks, \citep{DBLP:journals/corr/GideonKADP17} have exploited a progressive network originally proposed by \citep{DBLP:journals/corr/RusuRDSKKPH16} which was able to potentially support transferring knowledge across sequences of tasks. \citep{DBLP:journals/corr/GideonKADP17} have successfully transferred learning between three paralinguistic tasks: emotion, speaker, and gender recognition with an emphasis on speech emotion detection as the target application without the catastrophic forgetting effect. Their system outperformed the recent speech emotion recognition approaches utilizing fine-tuning pre-trained models and also performed significantly better than deep leaning models without the use of transfer learning techniques \citep{DBLP:journals/corr/RusuRDSKKPH16}. Nonetheless, an unavoidable limitation of this approach is that it is
computationally intensive since a number of new networks keep on growing according to the demand for a increasing number of new tasks which need to be
learned \citep{NIPS2017_7051}.
More recently, in order to alleviate the aforementioned downsides, \citep{DBLP:journals/corr/FernandoBBZHRPW17} have proposed PathNet as an alternative novel learning algorithm
for transfer learning. PathNet was designed as a neural network in which agents 
(e.g., pathways through different layers of the neural network) were embedded to discover which
parts of the network to be re-used for new tasks \citep{DBLP:journals/corr/FernandoBBZHRPW17}. Agents also hold an accountability for determining which subset of parameters to be used and to be updated for subsequent learning. Pathways through the neural network
for replication and mutation were selected by a tournament selection genetic algorithm proposed by \citep{Harvey:2009:MGA:2017762.2017781} during learning. The parameters along an optimal path evolved on the source task were fixed and a new population of
paths for the destination task is re-evolved. Such a manner of transfer learning
enables the destination task to be learned faster than learning from scratch or
fine-tuning, and therefore has greatly succeeded in several supervised learning
classification problems \citep{DBLP:journals/corr/FernandoBBZHRPW17}.

\begin{figure}[!t]
\centering
{\includegraphics[width=1.\linewidth]{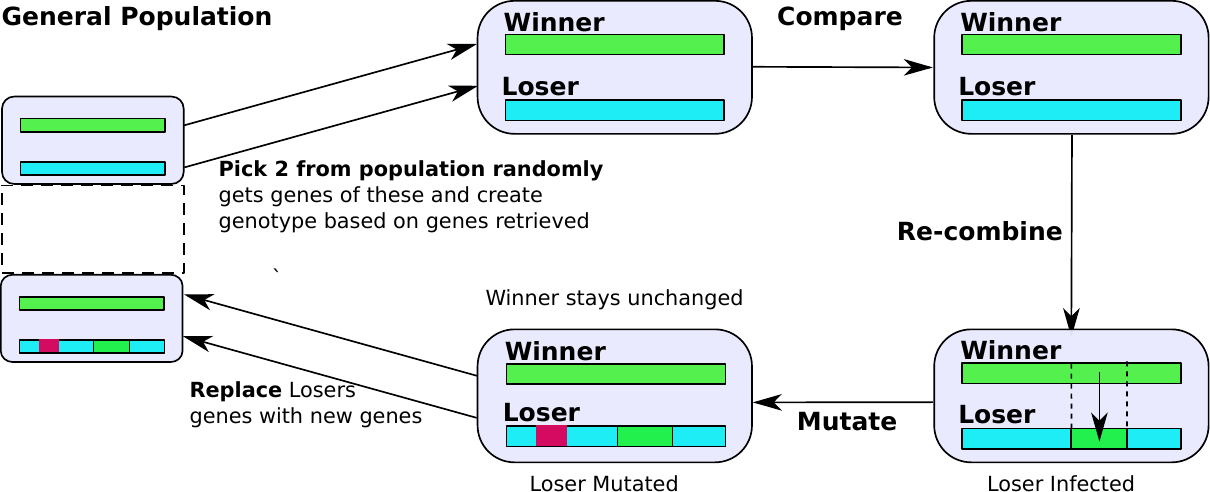}}
\caption{ The genotypes of the population are viewed as a pool of strings. One single cycle of the  Microbial  GA  is operated by initially \textbf{randomly picking} two, and subsequently \textbf{compare} their 
fitnesses  to  determine  \textbf{Winner}, \textbf{Loser},  and finally \textbf{recombine} where  some  proportion  of Winner\textquoteright s genetic material infects the \textbf{Loser}, before \textbf{mutating} the revised version of \textbf{Loser} (from \citep{Harvey:2009:MGA:2017762.2017781}).}
\label{Generic_Algorithm_Final}
\end{figure}

Motivated by the success of PathNet in those applications, in this paper, we explore utilizing PathNet for the facial expression based and speech based emotion recognition tasks. In this work, we first investigate how effective Path-Net is in transferring emotional knowledge from one visual emotion domain to another visual emotion domain to improve overall performance. Based on the experience we have gained in knowledge transferring ability of PahtNet within the visual domain in our previous work \citep{8545411}, we next investigate whether similar techniques can be used for transferring knowledge within the speech domain. Specifically we investigate the use of PathNet for speech emotion recognition (i) by exploring how well emotional knowledge learned from one speech emotion dataset could be transferred into another speech emotion dataset, and (ii) by examining how well emotional knowledge learned from multiple speech emotion datasets could be transferred to a single speech emotion dataset.

The contributions of our paper are as follows:
\begin{itemize}
    \item We introduce a novel transfer learning approach for the emotion recognition task by utilizing PathNet to deal with the problem of insufficient annotated emotion data as well to deal with the catastrophic forgetting issue commonly experienced with traditional transfer learning  techniques. 

    \item We confirm, through experimental results, that our proposed system has a significant potential to accurately detect emotions and demonstrates its substantial success in transferring learned knowledge between different emotion datasets, as well as in transferring learned emotional knowledge from multiple speech emotion datasets to a single speech emotion dataset.

    \item We conduct various sets of within-corpus on three commonly used bench-marking emotion datasets EMODB, eNTERFACE, and SAVEE and we show that the performance of our proposed transfer learning approach for emotion recognition exceeds recent state-of-the-art transfer learning schemes based on fine-tuning/pre-trained models.
\end{itemize}

The remainder of this paper is organized as follows: Section 2 describes related research; Section 3 presents our proposed system; Section 4 reports our
experimental results; and Section 5 concludes the paper.

\section{Related work}
A number of studies using transfer learning approaches have recently been proposed for the facial expression and speech emotion recognition task. Since the literature review for the facial expression recognition task utilizing deep learning and transfer learning method was reviewed and discussed in our previous work, the interested reader is referred to that work for detailed discussion and analysis, this section will only focus on reviewing the speech emotion recognition task.

\subsection{Speech Emotion Recognition using Deep Learning Techniques}

Deep learning techniques have emerged as powerful solutions in a wide variety of applications including natural language processing and computer vision owing to their inherent capability of directly learning a hierarchical feature representation from the input data \citep{Denet2015}. Inspired by their success in multiple fields, many deep learning approaches have been investigated for the task of speech emotion recognition. \citep{kim2017acmmm} have proposed a novel architecture for the speech emotion recognition task, in which long short-term memory (LSTM), fully convolutional neural network (FCN), and convolutional neural network (CNN) were combined aiming at extracting local invariant features from the spectral domain. By this combination, long-term dependencies have been well captured, thereby making utterance-level features more
discriminative \citep{kim2017acmmm}. Moreover, by embedding identity skip-connection techniques in their temporal architecture, this proposed system avoided the over-fitting problem caused by training on small amounts of data \citep{kim2017acmmm}. In another study, in order to handle the large mismatch between training and testing data, \citep{kim2017interspeech} have proposed utilizing multi-task learning, and then investigated gender and naturalness as auxiliary tasks of their proposed system. This can enhance significantly the capabilities of generalizing the speech emotion recognition models. The experimental results evaluated on within-corpus and cross-corpus scenarios have shown good performance of their proposed system.

In other approaches, \citep{7926723,NGUYEN2018} have proposed the learning of spatio-temporal features with C3Ds from audio and video for multimodal emotion recognition. \citep{kim2017acii} have also proposed three dimensional convolutional neural networks (C3Ds) to address the challenge of modeling the spectro-temporal dynamics for speech emotion recognition by simultaneously extracting  short-term and long-term spectral features with a moderate number
of parameters. \citep{10995} have exploited the adversarial auto-encoders focusing on (i) compressing high dimensional vectors encoding emotional utterances into low space vectors (referred to as code vectors) without sacrificing the discrimination during classifying the original vectors, and (ii) generating synthetic samples applying the adversarial auto-encoder, subsequently used for emotion classification. Their system has mainly concentrated on detecting emotions at utterance level features instead of frame-level ones. \citep{7883728} have proposed a speech emotion recognition system, in which spectrograms were initially extracted from speech signals at different frequencies. Such spectrograms were subsequently fed into a CNN for emotion prediction. Similarly, \citep{7952656} have addressed emotional valence in human speech by directly learning spectrograms of emotional speech using a CNN. In order to further improve the performance, this architecture has been extended to a deep convolutional generative neural network which was trained in an unsupervised fashion.

Researchers have also explored the whispered speech emotion recognition task,  where different feature transfer learning approaches have been explored by utilizing shared-hidden-layer auto-encoders, extreme learning machines auto-encoders, and denoising auto-encoders \citep{7879177}. The key ideas of these approaches were to develop a transformation for automatically capturing useful features hidden in data and to transfer the knowledge from the target domain-testing (whispered speech) to the source domain-training (normal phonated speech), consequently leading the great benefit regarding optimizing all parameters with the support from the test set. Extensive experiments have been conducted with a focus on entirely tackling the binary classification (i.e. valence/arousal) \citep{7879177}. In another study, \citep{7862157} have also pointed out that many speech emotion recognition systems usually demonstrate poor performance on speech data when there is significant differences between training and test speech arising from the variations in the linguistic content, speaker accents, and domain/environmental conditions \citep{7862157}. To further improve such systems performing under the mismatched training and testing condition, a novel unsupervised domain adaptation algorithm has been introduced and trained by simultaneously learning discriminative information from labeled data and incorporating the prior knowledge from unlabeled data into the learning.

\subsection{Speech Emotion Recognition using Transfer Learning Techniques}

However, recent studies into emotion recognition have been hindered by the lack of large databases for learning \citep{7956190,KAYA201766}. To address the lack of large emotion datasets, the fine-tuning/pre-trained model has been recently widely investigated for the emotion recognition task \citep{KAYA201766,8085174,kim2017acmmm,kim2017acii,kim2017interspeech,DBLP:journals/corr/abs-1801-06353} in which the CNN architectures were pre-trained using the generic ImageNet dataset and fine-tuned on emotion datasets \citep{Ng:2015:DLE:2818346.2830593,KAYA201766}. To learn audio features, \citep{7956190} used a pre-trained C3D models on large-scale image and video classification datasets, and then fine-tuned them on emotion recognition tasks. To improve the performance of a speech emotion recognition system on such challenging conditions as cross-corpus and cross-language scenario, \citep{DBLP:journals/corr/abs-1801-06353} have proposed a transfer learning technique using deep belief networks (DBNs). The experimental results evaluated on five different corpora in three different languages demonstrated the robustness of their system. These results also indicated that use of a large number of languages and a small part of the target data during training could dramatically strengthen the emotion recognition accuracy. However, since they have attempted to validate the system on five different datasets annotated differently, their system only focused on addressing the classification for binary positive/negative valence.

In more recent studies, \citep{8085174} have reconfirmed that the low-level hand-engineered features seem not to be discriminative enough to recognize the subjective emotions \citep{8085174}. To address this disadvantage, \citep{8085174} have proposed to extract three channels of log Mel-spectrograms (static, delta, and delta delta corresponding to red, green, and blue in the RGB model of images) from segments over all utterances, and the pre-trained AlexNet model was then fine-tuned on those extracted features. A discriminant temporal pyramid matching technique was subsequently combined with optimal Lp-norm pooling, before exploiting a linear support vector machine to classify the final speech emotion score. Although this architecture demonstrated sufficient performance on tackling discrete speech emotion recognition, the system was unable to address the continuous emotion recognition task  \citep{8085174}.  \citep{nguyen2020joint} have proposed a joint deep cross-domain transfer learning for emotion recognition which was able to effectively jointly transfer the knowledge learned from rich datasets to source-poor datasets. Moreover, as discussed earlier, all of these fine-tuning approach based systems \citep{8085174,7956190,Ng:2015:DLE:2818346.2830593,KAYA201766} still have the drawbacks previously alluded to such as discarding previously learned information which were detailed by \citep{DBLP:journals/corr/RusuRDSKKPH16}. To mitigate these drawbacks, \citep{DBLP:journals/corr/GideonKADP17} have introduced a learning algorithm using the progressive networks proposed by \citep{DBLP:journals/corr/RusuRDSKKPH16} to effectively transfer knowledge captured from one emotion dataset into another. Although they have handled somewhat successfully the above mentioned limitations, their expensive computation, which kept on increasing when adding new tasks to be learned, makes them less applicable in the implementation of emotion recognition for real world applications.

\begin{figure*}[t!]
\centering
\subfloat[Video pre-processing steps. All frames are initially extracted from all videos, then face regions are detected by applying an improved Viola-Jones algorithm \citep{NGUYEN2018}, before being resized to 64$\times$64$\times$3.]{\includegraphics[width=0.8\linewidth]{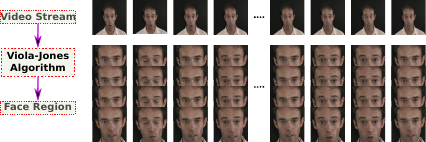}
\label{video_pre_processing.pdf}}
\hfil
\subfloat[Audio pre-processing steps. First, mel-spectrograms from segments over all utterances are extracted. Then, 3 channels of Mel-spectrograms with size 64 $\times$ 64 $\times$ 3 ($F$ = 64, $T$ = 64, $C$ = 3) corresponding to the static, delta, and delta-delta coefficients are extracted]{\includegraphics[width=0.8\linewidth]{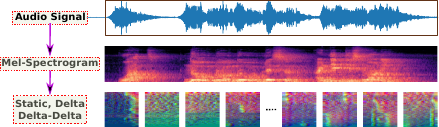}
\label{audio_pre_processing.pdf}}
\hfil
\caption{This figure illustrates the pre-processing steps for video and audio stream.}  
\label{Pre_processing}
\end{figure*}

\section{Proposed Methodology}
Our main goal in this paper is to investigate techniques to improve the accuracy of deep learning based emotion recognition task which is hindered by the non availability of large annotated emotion datasets. We achieve this goal by using the recently proposed innovation known as PathNet for transfer learning. We propose the use of a novel transfer learning approach by adopting PathNet to solve the facial expression emotion recognition, and the speech emotion recognition task. Our proposed system is illustrated by simple block diagram consisting of two main blocks which are an input pre-processing block (for video and speech) followed by our proposed PathNet block along with the output classifying block (as illustrated in Fig. \ref{Pathnet_architecture__}). The video and audio stream are initially pre-processed. For video stream, we initially exploit a Viola Jones-based algorithm \citep{7926723} to extract all face regions from both SAVEE and eNTERFACE \citep{1623803} datasets. This is described in detail in Section \ref{video_pre_processing}. For audio stream, we also initially extract three channels of log Mel-spectrograms (static, delta, and delta delta corresponding to red, green, and blue) from segments over all utterances from eNTERFACE \citep{1623803}, SAVEE \citep{HaqJackson_MachineAudition10}, and EMO-DB \citep{Burkhardt05adatabase} and these steps are described in detail in Section \ref{audio_pre_processing}. These extracted video and audio features are subsequently fed into our PathNet to classify a final facial expression and speech emotion score, respectively. To the best of our knowledge, the use of PathNet has not been previously investigated in dealing with the dearth of suitable emotion databases for the development of emotion recognition system. The procedures of feature extraction and our PathNet architecture are described in more detail in the following subsections.


\subsection{Video pre-processing}
\label{video_pre_processing}
All frames are initially extracted from visual signal for further steps. Since such extracted frames still contain considerable redundant information for emotion detection, we extract only the face regions using the simple algorithm \citep{7926723} as follows:
\begin{enumerate}
\item All bounding boxes containing face regions in each frame were extracted employing the Viola-Jones algorithm \citep{Viola:2004:RRF:966432.966458} and a face region was then detected. 
\item In some cases where the Viola Jones algorithm detected no faces, or more than 1 face, the location of the previously detected face region was used.
\end{enumerate}
By applying this algorithm, we have successfully extracted all face regions from all frames in both datasets (SAVEE and eNTERFACE) as input into PathNet (see in Fig. \ref{Pre_processing} (a). as an illustration of some input samples)

\subsection{Audio pre-processing}
\label{audio_pre_processing}

\citep{8085174} have pointed out that the low-level hand-engineered features, which includes RASTA-PLP \citep{Hermansky:1992:RSA:1895550.1895585}, pitch frequency features, energy-related features \citep{1326055}, formant frequency \citep{1577304}, Zero Crossing Rate (ZCR) \citep{6778857}, Mel-Frequency Cepstrum Coefficients (MFCC) and its first derivative, Linear Prediction Cepstrum Coefficients (LPCC), Linear Prediction Coefficients (LPC) \citep{1699080,6023178}, seem not to be discriminative enough for recognizing the subjective emotions \citep{8085174}. Therefore, in order to boost the performance of speech emotion recognition system, instead of exploiting such hand-crafted features, \citep{8085174} have extracted three channels of log Mel-spectrograms from segments over all utterances, and then fine-tuned the pre-trained AlexNet model on these extracted features. Their proposed architecture has successfully addressed the speech emotion recognition task. Motivated by this high performance, in this paper we also propose to extract these features as follows (see Fig. \ref{Pre_processing} (b)):
\begin{itemize}
    \item Mel-spectrogram segments with size 64 $\times$ 64 $\times$ 3 ($F$ = 64, $T$ = 64, $C$ = 3) are generated from 1-D speech signals. $F$ denotes as the number of Mel-filter banks, $T$ denotes as the segment length corresponding to the frame number in a context window, and C represents the number of channels of Mel-spectrogram. The 3 channels of Mel-spectrograms are the static, delta and, delta-delta coefficients of Mel-spectrogram), respectively \citep{8085174}.
\item Specifically, 64 Mel-filter banks from 20 to 8000 Hz are exploited to extract the whole log Mel-spectrogram using a 25ms Hamming window size with 10ms overlapping for an utterance. A context window of 64 frames (its length 10ms $\times$ 63 + 25ms = 655ms) is then applied to the whole log Mel-spectrogram to extract the static 2-D Mel-spectrogram segments (64 $\times$\ 64) with overlapping size of 30 frames \citep{8085174}.

\end{itemize}

\begin{figure*}[t!]
\centering
\subfloat[Our proposed emotion recognition block diagram]{\includegraphics[width=1.\linewidth]{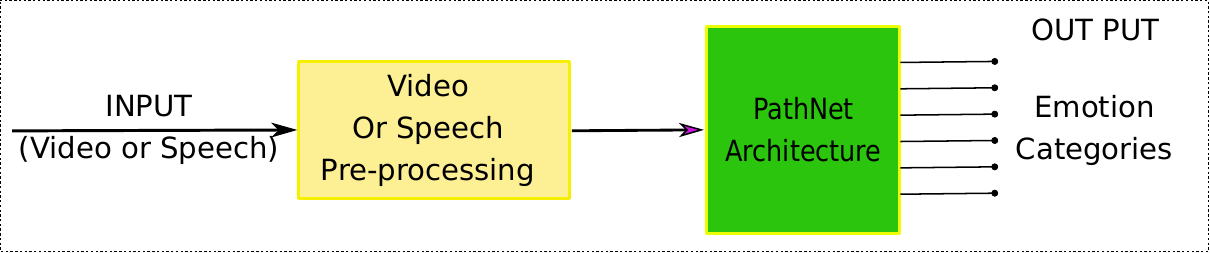}
\label{Pathnet_architecture_}}

\subfloat[The PathNet architecture has 3 layers, and each layer consists of 20 modules, and each module is followed by a rectified linear unit. The feature maps between layers are averaged before being fed into the modules of the next layer. A maximum of 4 modules per layer are typically allowed to be presented in a pathway.]{\includegraphics[width=1.\linewidth]{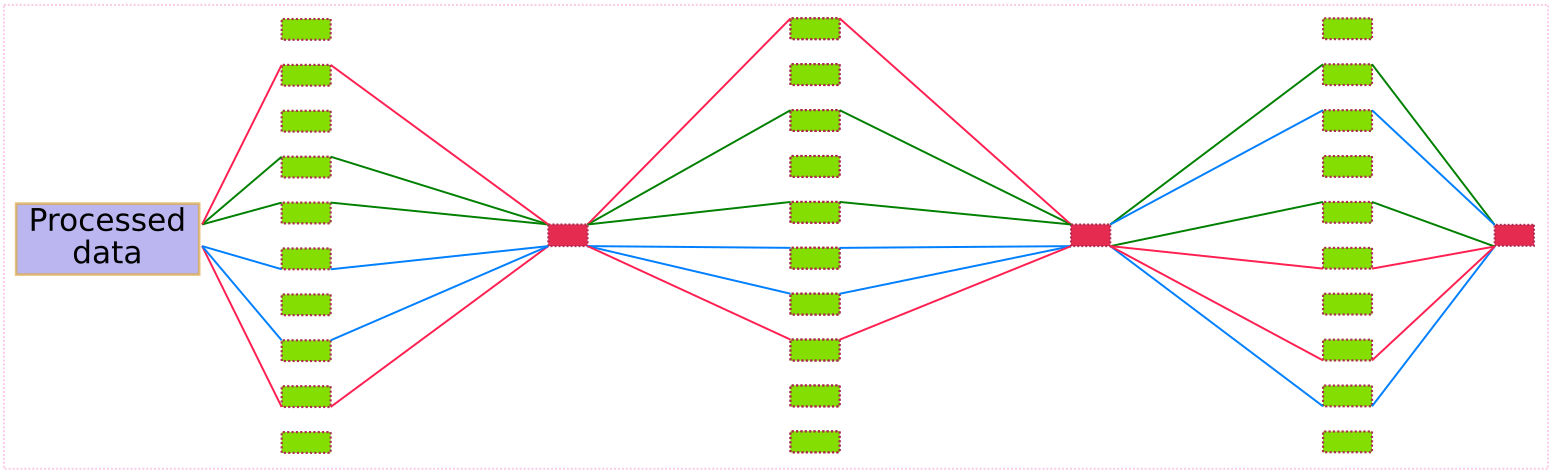}
\label{Pathnet_architecture}}

\subfloat[Description of each component of our PathNet architecture]{\includegraphics[width=1.\linewidth]{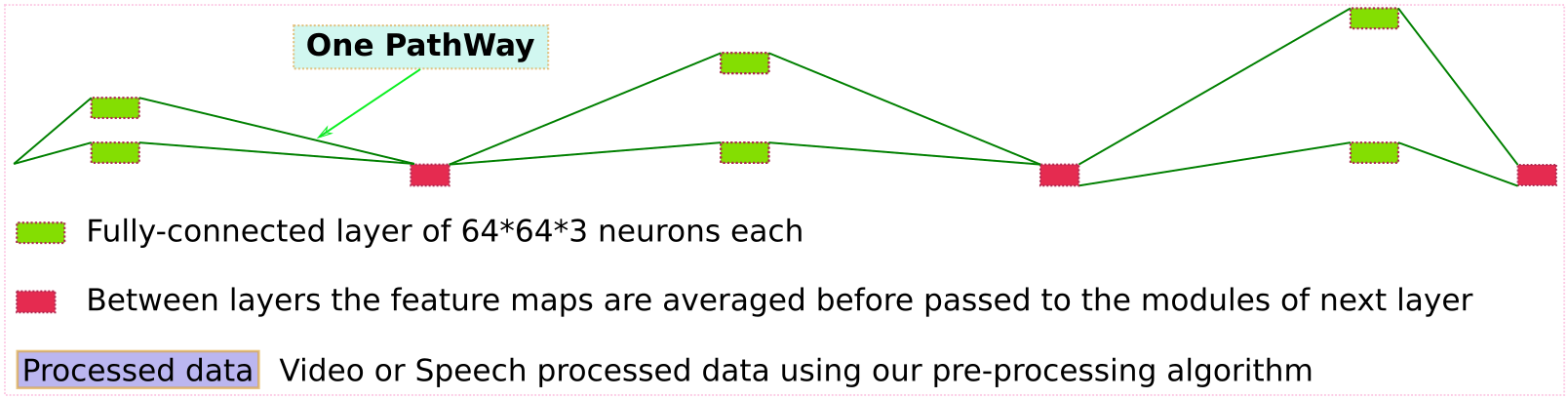}
\label{Components_of_pathnet}}
\hfil
\caption{Illustrates our proposed emotion recognition system}  
\label{Pathnet_architecture__}
\end{figure*}

\subsection{PathNets}

\label{PathNets}
Our PathNet architecture and its settings relies on PathNets \citep{DBLP:journals/corr/FernandoBBZHRPW17} which was used to conduct all sets of experiments on CIFAR \citep{Krizhevsky09} and SVHN \citep{37648}. The following sections will provide in detail its architecture and explain further how to train our system, how to transfer learned emotional knowledge between emotion dataset.

\subsubsection{PathNet architecture}

Our PathNets includes a number of layers (L = 3), a number of modules (M = 20) per layer of 20 neurons in each. Each module itself functions as a neural  network consisting of linear units, and followed by a transfer function (rectified linear units adopted). For  each  layer  the  outputs  of  the  modules of this layer are averaged before being fed into the active modules of the subsequent layer.  A module is active if it is shown in the path genotype and currently validated (shown in Fig. \ref{Pathnet_architecture__} (b)). A maximum of 4 distinct modules per  layer are typically allowed in a pathway. The final layer is not shared  for  each  task  which is being learned \citep{DBLP:journals/corr/FernandoBBZHRPW17}. 

\subsubsection{Pathway Evolution and Transfer Learning Approach}
One emotion dataset (source data) is trained for a fixed number of generations with a goal of finding an optimal pathway by adopting a binary  tournament selection algorithm proposed by \citep{Harvey:2009:MGA:2017762.2017781} (see Fig. \ref{Generic_Algorithm_Final}) which takes responsibility for eliminating bad configurations and mutating good ones, and subsequently training them further. This pathway is then fixed. This means that its  parameters are no longer  permitted to modify and the rest of parameters, which are not shown in such best fit path, are  reinitialized, and  are then again trained/evolved on the another emotion dataset (destination data). Through this knowledge transferring approach, the destination data is permitted to be learned faster than learning from scratch or after fine-tuning. The performance measurement of our proposed system is the recognition accuracy achieved after such fixed training time. Evidence to confirm a positive transfer in these cases is given by a better final recognition accuracy of our proposed system achieved when trained on destination data than  that achieved by learning from scratch.

\begin{figure}[!t]
\centering
{\includegraphics[width=1.\linewidth]{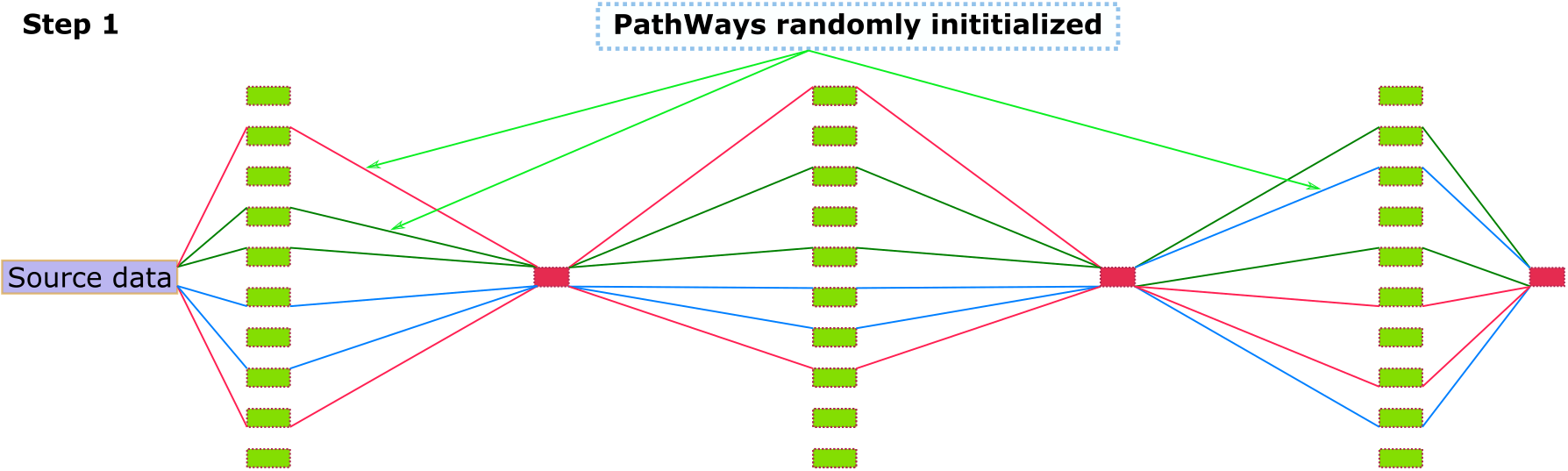}}
\caption{A number of pathways is randomly initialized when pathnet learning on source data}
\label{Step1.PNG}
\end{figure}

\begin{figure}[!t]
\centering
{\includegraphics[width=1.\linewidth]{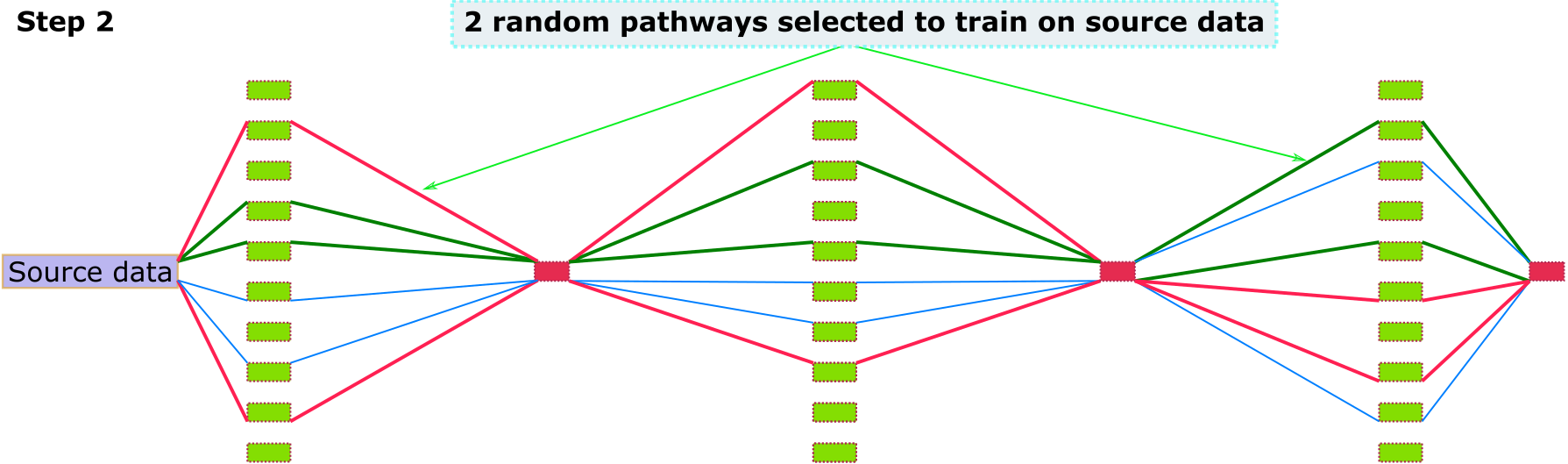}}
\caption{Two pathways are selected to train on source data}
\label{Step2.PNG}
\end{figure}

\begin{figure}[!t]
\centering
{\includegraphics[width=1.\linewidth]{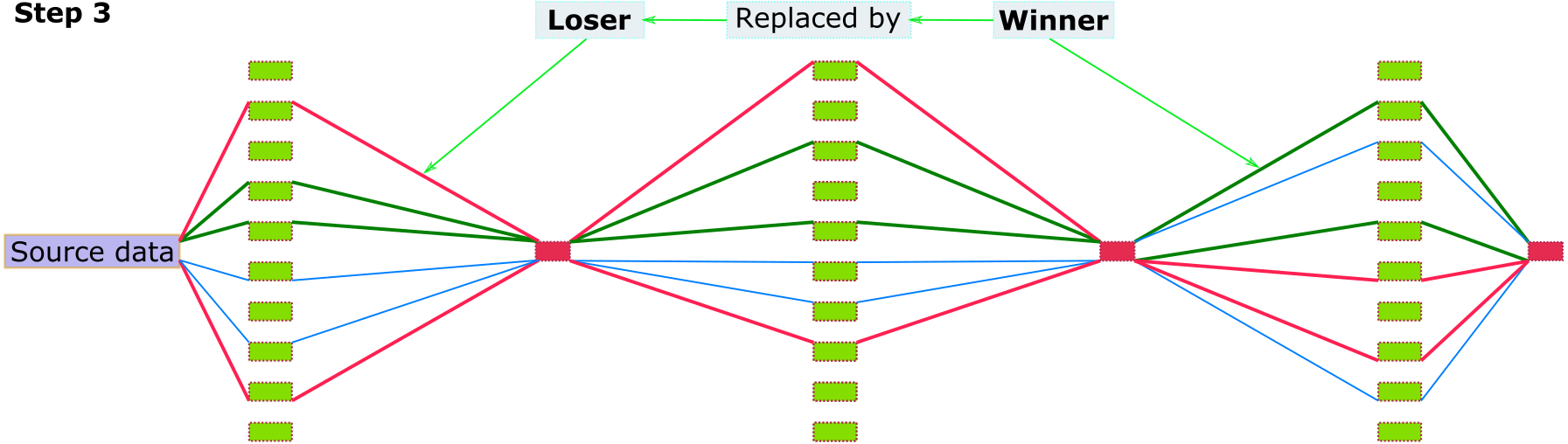}}
\caption{A pathway with bad performance (loser) is replaced by a pathway with better performance (winner)}
\label{Step3.PNG}
\end{figure}

\begin{figure}[!t]
\centering
{\includegraphics[width=1.\linewidth]{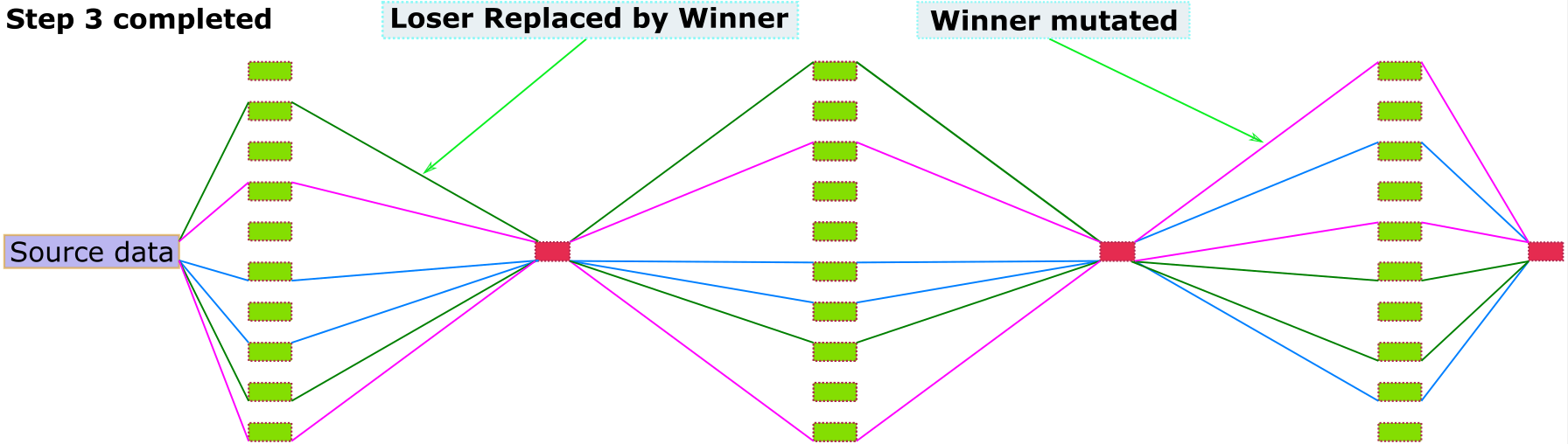}}
\caption{Winner is mutated}
\label{After_Step3.PNG}
\end{figure}

\begin{figure}[!t]
\centering
{\includegraphics[width=1.\linewidth]{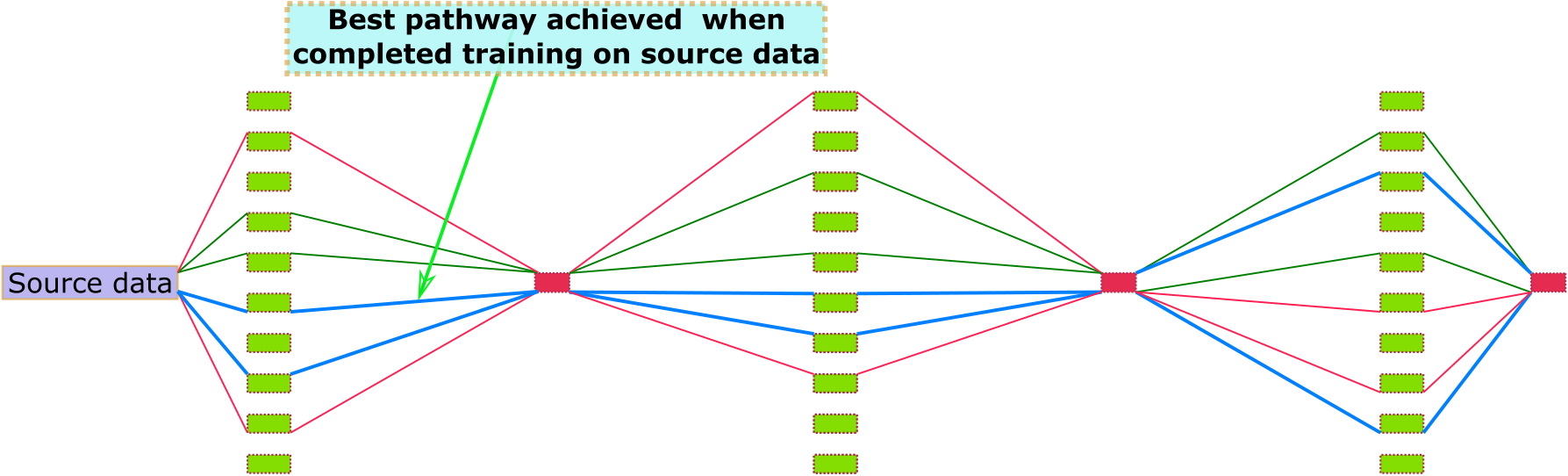}}
\caption{Best pathway achieved when completed training on the source data}
\label{Best_Sour.PNG}
\end{figure}

When PathNet is trained on source data, at the beginning, a population of genotypes is randomly generated (See Fig. \ref{Step1.PNG}). In each generation, two pathways are randomly selected to train on source data (see Fig. \ref{Step2.PNG}). The reason only two pathways are selected is that binary tournament selection is exploited to choose the pathways. One pathway is learned using stochastic gradient descent for $\small{T}$ epochs ($\small{T}$ equals the number of samples counted in training set divided by mini-batch size). Another pathway is also learned using stochastic gradient descent for $\small{T}$ epochs ($\small{T}$ equals the number of samples counted in training set divided by mini-batch size). The fitness of such pathways is the rate of correct samples on the training set during that period of training time. When training two pathways is completed, the pathway with the bad performance (called loser) is replaced by the pathway with the better performance (called winner) and the winner is then mutated  with equal probability 1/(4$\times$3) per
each candidate of the genotype (see Fig. \ref{Step3.PNG}, and Fig. \ref{After_Step3.PNG}), a new random integer from range [-2, 2] is added to the current value of the winner candidate. We repeat step 2 (Fig. \ref{Step2.PNG}) and step 3 (Fig. \ref{Step3.PNG}) for number of generations. When training on source data is completed we achieve best pathway (see Fig. \ref{Best_Sour.PNG}).

\begin{figure}[!t]
\centering
{\includegraphics[width=1.\linewidth]{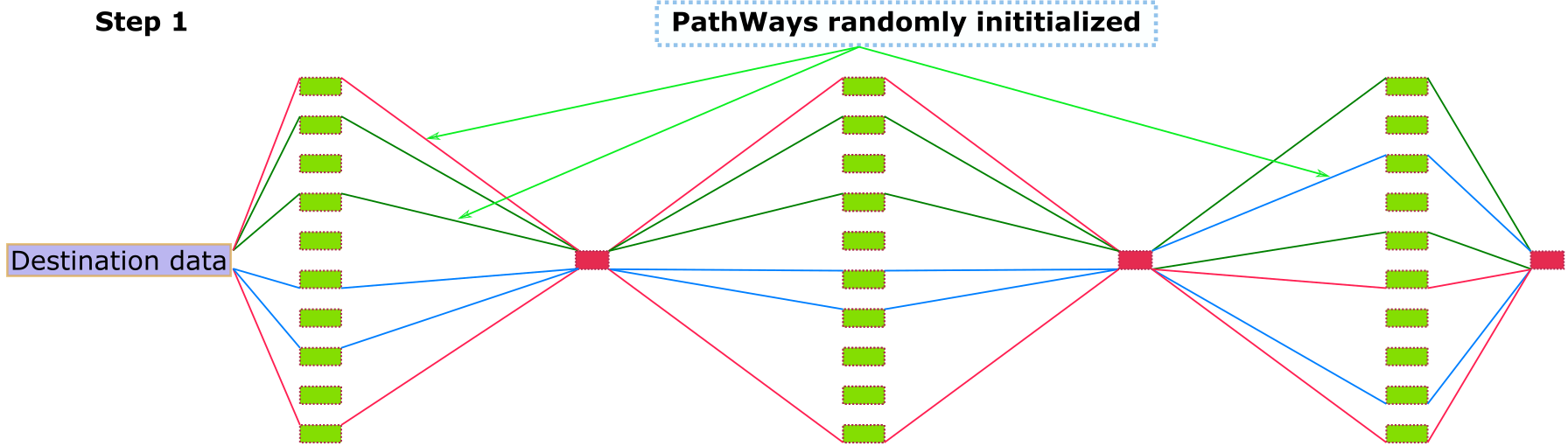}}
\caption{A number of pathways are randomly initialized when pathnet trained on the destination data}
\label{Step1_Des.PNG}
\end{figure}

\begin{figure}[!t]
\centering
{\includegraphics[width=1.\linewidth]{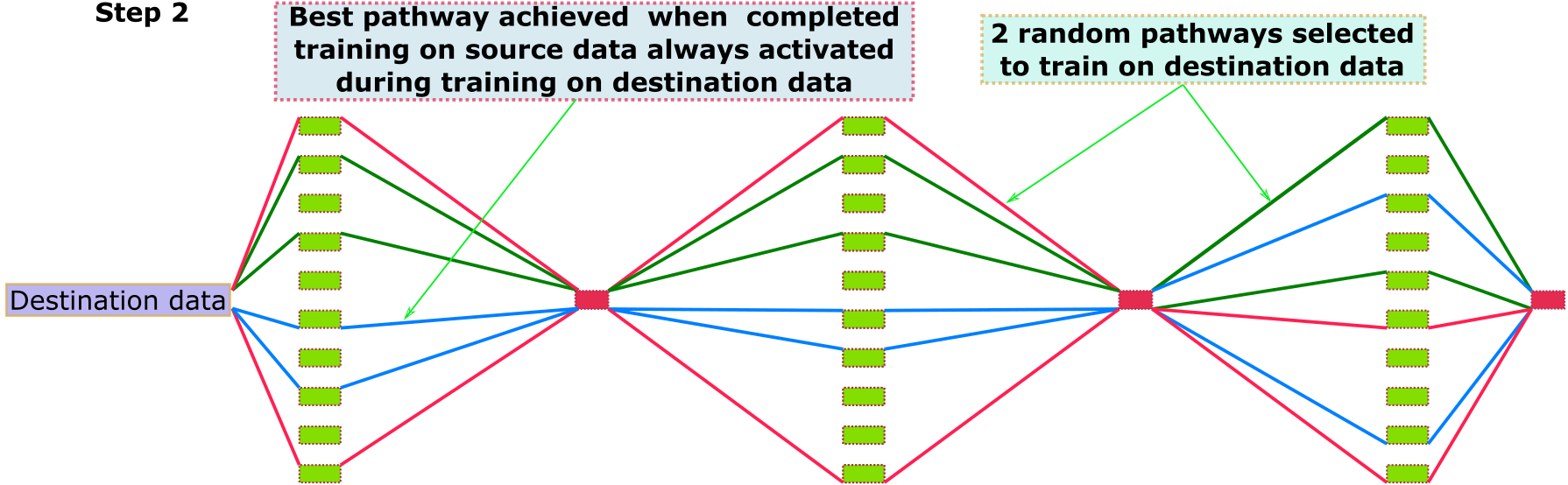}}
\caption{Two random pathways are also selected to train on the destination data}
\label{Step2_Des.PNG}
\end{figure}
\begin{figure}[!t]
\centering
{\includegraphics[width=1.\linewidth]{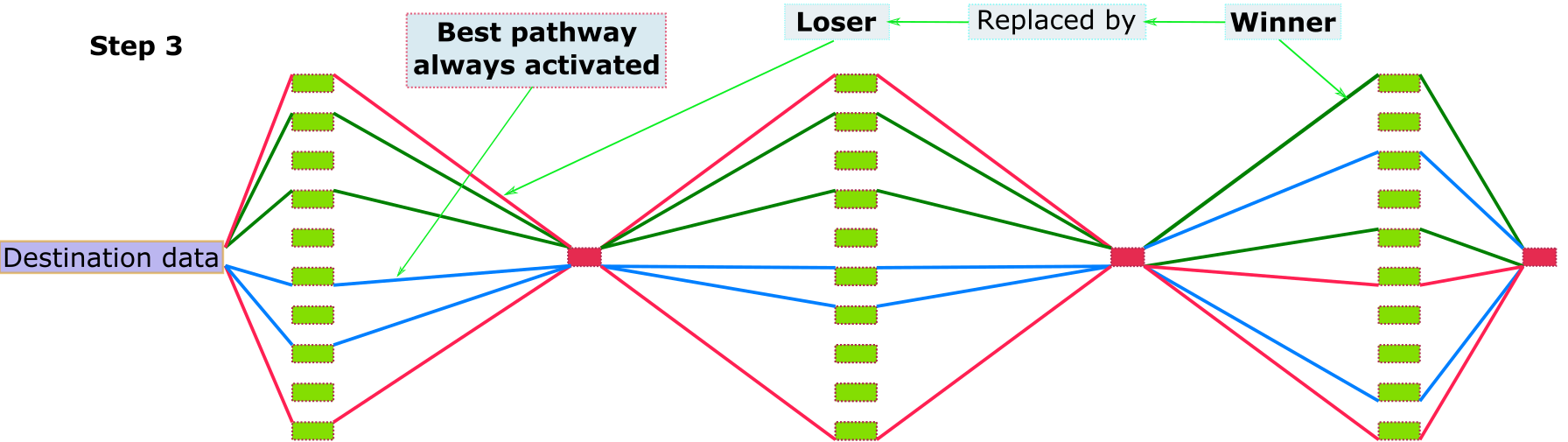}}
\caption{A pathway with bad performance (loser) is replaced by a pathway with better performance (winner) when learning with the destination data}
\label{Step3_Des.PNG}
\end{figure}

The parameters presented in this best fit pathway are fixed and are reused for training on the destination data and the rest of parameters are randomly reinitialized. When training PathNet on the destination data, the procedures for this training stage are similar to ones when training on the source data (procedures illustrated in Fig. \ref{Step1_Des.PNG}, Fig. \ref{Step2_Des.PNG}, and Fig. \ref{Step3_Des.PNG}). At the beginning, a new population of genotypes is randomly initialized and then trained/evolved further on the destination data. However, the difference is that the best pathway achieved when completed training on source data is always activated during learning with the destination data. When completed training on the destination data, we also achieve best pathway (see Fig. \ref{Best_Des.PNG}).

It is quite difficult to fully understand how to train/test PathNet and how it works by only reading textual explanation that is presented in this section. Therefore, in order to make our PathNet based transfer learning approach more understandable, we have added further explanation on the progression of achieving the best pathway by visualizing every step using the corresponding figure (please see Figs. 3.3 - 3.13). As we can see these Figs, to ease for readers to view and follow step by step, we simplify the architecture by drawing only ten modules in each layers, up to two modules are activated in each layer, and a population of four random pathways are initialized. Whereas the Figs. 3.19- 3.23 are drawn to further explain and visualize the progression of how to achieve best pathway for one particular set of experiment. These Figs illustrate exact parameters and architectures corresponding to the progression in gaining the optimal pathway when conducting the set of experiment. We believe that this explanation is more comprehensive than done by the original PathNet paper \citep{DBLP:journals/corr/FernandoBBZHRPW17} as in that paper authors did not illustrate such steps using figures.

\begin{figure}[!t]
\centering
{\includegraphics[width=1.\linewidth]{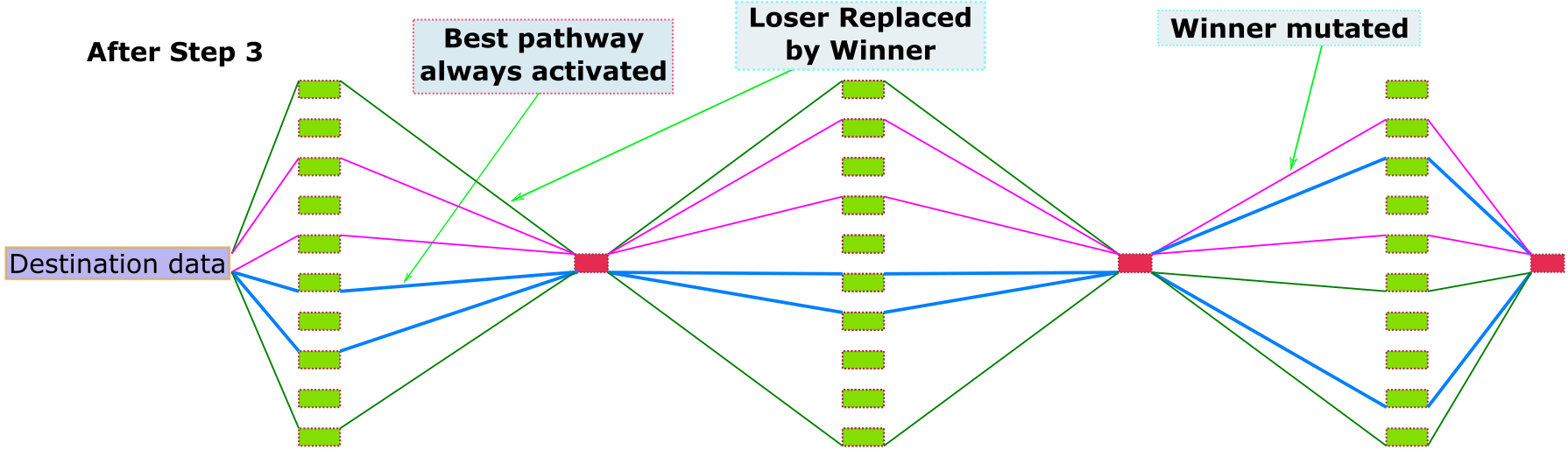}}
\caption{The winner is then mutated}
\label{After_Step3_Des.PNG}
\end{figure}

\begin{figure}[!t]
\centering
{\includegraphics[width=1.\linewidth]{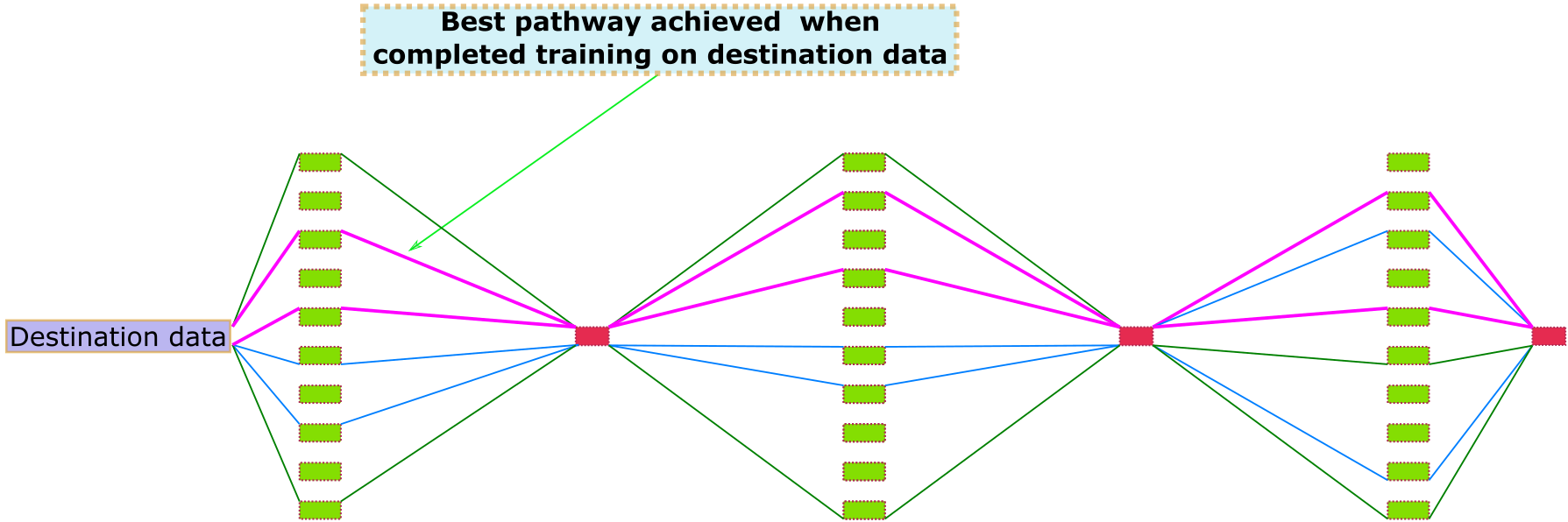}}
\caption{Best pathway is achieved when completed training on the destination data}
\label{Best_Des.PNG}
\end{figure}

\section{Experiments \& Results}
 \textbf{Dataset Details:} The eNTERFACE dataset \citep{1623803} is an audio-visual dataset which has 44 subjects and includes a total of 1293 video sequences in which the proportion of sequences corresponding to women and men are 23\% and 77\%, respectively. They were asked to express 6 discrete emotions including anger, disgust, fear, happiness, sadness, and surprise \citep{1623803}.

The SAVEE dataset \citep{HaqJackson_MachineAudition10} is an audio-visual dataset which was recorded by higher degree researchers (aged from 27 to 31 years) at the University of Surrey, and four native male British speakers. All of them were also required to speak and express seven discrete emotions such as anger, disgust, fear, happiness, sadness, surprise, and neutral. The dataset comprises of 120 utterances per speaker, resulting in a total of 480 sentences \citep{HaqJackson_MachineAudition10}.

The EMO-DB dataset \citep{Burkhardt05adatabase} is an acted speech corpus  containing 535 emotional utterances with seven different acted emotions listed as disgust, anger, neutral, sadness, boredom, and fear. These emotions were stimulated by five male and five female professional native German-speaking actors, generating five long and five short sentences German utterances used in daily communication. These actors were asked to read predefined sentences in the targeted seven emotions. The audio files are on average around 3 seconds long were recorded using an anechoic chamber with high-quality recording equipment at a sampling rate of 16 kHz with a 16-bit resolution and mono channel \citep{8085174}.

\textbf{Performance Measures:} We evaluate our proposed system on all three datasets: the SAVEE, the eNTERFACE, and the EMO-DB dataset.  We apply k-fold cross-validation, the original training data is randomly divided into k equal parts. Of the k-parts, one of them is fixed as the validation data for testing the model, and the other k-1 parts are used as training data. The cross-validation process is then repeated 5 times. We also apply the leave-one-subject-out cross-validation protocol which means that we have to conduct N experiments if the dataset consists of N subjects. For each experiment, N-1 subjects are included in the training set and the remaining subject is used for the testing set and all of our sets of experiments are carried out in a subject independent manner. 

Apart from these, Weighted Averaged Recall (WAR) has been recently adopted and has become widely a standard measure for evaluating the performance of emotion recognition systems \citep{8085174}. In addition, Unweighted Averaged Recall (UAR) \citep{7160715,eyben2015real,8085174} has been also popularly adopted to evaluate this performance with respect to reflecting unbalance between emotional classes \citep{8085174}. Therefore, in order to fairly compare with some recent state-of-the-art speech emotion recognition schemes using the same above-mentioned measures (WAR, UAR), in this paper, we also compute and compare both Weighted Averaged Recall and Unweighted Averaged Recall to validate the performance of our proposed system. Table \ref{evaluation measure} illustrates the detail formulas on how to calculate these evaluating measures.

\newcolumntype{C}{>{\centering\arraybackslash}m{0.49\textwidth}}
\newcolumntype{D}{>{\raggedleft\arraybackslash}m{0.12\textwidth}}
\newcolumntype{K}{>{\raggedleft}m{0.28\textwidth}}
\newcolumntype{L}{>{\centering}m{0.26\textwidth}}
\newcolumntype{E}{>{\centering}p{0.12\textwidth}}

\begin{table}[t]
\begin{center}
\caption{Different measures for multi-class classification $C_i$. For class $C_i$$, tp_i$ are true positive, and $fp_i$ - false positive, $fn_i$ - false negative, and $tn_i$ - true negative counts, $l$ - the number of classes, \citep{SOKOLOVA2009427}.} 

\begin{tabular}{LDC}
\midrule
 \textbf{Measure}&\textbf{Formula}& \textbf{Evaluation focus} \\
\midrule

Unweighted Averaged Precision (UAP)&$
    \frac{\sum_{i=1}^{l}\frac{tp_i}{tp_i+fp_i}}{l}$&An average per-class agreement of the data class labels with those of a classifiers\\
\midrule
Unweighted Averaged Recall (UAR)& $\frac{\sum_{i=1}^{l}\frac{tp_i}{tp_i+fp_i}}{l}$&An average per-class effectiveness of a classifier to identify class labels\\

\midrule
\end{tabular}
\label{evaluation measure}
 
\end{center}
\end{table}

\newcolumntype{B}{>{\centering\arraybackslash}m{0.28\textwidth}}
\newcolumntype{A}{>{\centering\arraybackslash}m{0.64\textwidth}}
\newcolumntype{O}{>{\raggedleft}m{0.28\textwidth}}
\newcolumntype{M}{>{\centering}m{0.3\textwidth}}

\begin{table}[t]
\begin{center}
\caption{Description of notation of all models for all sets of experiments of our proposed system conducted on eNTERFACE, SAVEE, and EMODB;}

\begin{tabular}{A|B}

\hline
  \textbf{Transferring emotional knowledge}&\textbf{Notation}\\
\hline
\hline
 PathNet is trained from scratch on visual eNTERFACE&V\textunderscore{eNTER}\\
\hline
PathNet is trained from scratch on visual SAVEE&V\textunderscore{\text{SAV}}\\
\hline
PathNet is trained from scratch on audio SAVEE&A\textunderscore{\text{SAV}}\\
\hline
PathNet is trained from scratch on audio EMODB&A\textunderscore{\text{EMO}}\\
\hline
From visual eNTERFACE to visual SAVEE&V\textunderscore{\text{eNTER}$\rightarrow$ \text{SAV}}\\
\hline

From visual SAVEE to visual eNTERFACE&V\textunderscore{\text{SAV}$\rightarrow$ \text{eNTER}}\\
\hline
From audio eNTERFACE to audio SAVEE &A\textunderscore{\text{eNTER}$\rightarrow$ \text{SAV}}\\
\hline
From audio eNTERFACE and audio SAVEE to audio EMODB&A\textunderscore{\text{eNTER+SAV}$\rightarrow$ \text{EMO}}\\

\hline
\end{tabular}
\label{list_of_experiment}
 
\end{center}
\end{table}

\begin{table}[!t]
\begin{center}
\caption{Results of our proposed system evaluated on visual SAVEE, visual eNTERFACE, audio SAVEE, and audio EMODB. In the first two rows, we explore transferring the emotional knowledge from visual eNTERFACE to visual SAVEE and vice versa. In the last two rows, the emotional knowledge is transferred from audio eNTERFACE to audio SAVEE. We also explore transferring the emotional knowledge from multiple audio emotion domains (audio eNTERFACE and audio SAVEE) to one audio emotion domain (audio EMODB).}
\label{cross-corpus experimental results}
\begin{tabular}{c||c|c|c|c|c|c||c}
\hline
\textbf{Method} & \textbf{Ang} & \textbf{Sur} & \textbf{Dis} & \textbf{Fea} & \textbf{Hap} & \textbf{Sad}& \textbf{WAR}\\
\hline\hline
\textbf{V$_{\text{eNTER}\rightarrow \text{SAV}}$}&0.93&0.94&0.96&0.92&0.94&0.95&0.94\\
\hline
\textbf{V$_{\text{SAV}\rightarrow \text{eNTER}}$} &0.99&1.0&0.95&0.81&1.0&0.93&0.94\\
\hline
\textbf{A$_{\text{eNTER}\rightarrow \text{SAV}}$}& 0.73&0.70 &0.56 &0.72 &0.80 &0.77&0.71\\
\hline
\textbf{A$_{\text{eNTER+SAV}\rightarrow \text{EMO}}$}& 0.99&0.99 &1.0 &0.94 &0.94 &0.96 &0.97\\
\hline
\end{tabular}
\end{center}

\end{table}

As our baseline systems, we use the methodology proposed by \citep{8085174} and a off-the-shelf CNN (AlexNet) as our baseline systems. Additionally, we also implement additional baseline systems and each of which is trained on the same emotion dataset from scratch using PathNet.

Since our main focus of this paper is to address the issue of lack of emotion data for deep learning techniques. We clearly introduced this problem in the \textbf{Abstract} and further discussed it in the \textbf{Introduction} section. To make compatible with the problem we are trying to solve, we should choose small emotion datasets to show that our proposed methodology demonstrates a robust performance on such poor data. Furthermore, all large-scale datasets such as CK+, RAF-DB and AffectNet only can be used for facial expression recognition which is not the main task of this paper. It is, however, noteworthy that addressing the issue of insufficient data for speech emotion recognition task is the main work of our paper.



We conduct various sets of experiments using our proposed system in examining how well the accuracy of facial expression and speech emotion recognition are improved when transferring learned emotional knowledge between emotion datasets. In the first two sets of experiments, we conduct experiments for facial expression recognition, in which we transfer emotional knowledge from visual emotion eNTERFACE dataset (source data) to visual emotion SAVEE dataset (destination data) and vice versa. For the next two sets of experiments, we conduct experiments for speech emotion recognition by transferring emotional knowledge from audio emotion eNTERFACE dataset (source data) to audio emotion SAVEE dataset (destination data), and transferring emotional knowledge from multiple audio emotion datasets (eNTERFACE and SAVEE) to audio emotion EMODB.

\textbf{PathNet Settings:} Our proposed PathNet architecture (i.e. $\small{L} = 3$ layers, $\small{M}$ = 20 liner units per layer of 20 neurons each followed by rectified linear units, average function used to activate units between two layers, and a maximum of 4 of those units per layer included in a pathway which is represented by a 4$\times$3 matrix of intergers in the range [1,13]). Our proposed PathNet architecture is trained on visual eNTERFACE/visual SAVEE/audio eNTERFACE/(audio eNTERFACE and audio SAVEE) which are set as as source data and is then trained/evolved on visual SAVEE/visual eNTERFACE/audio SAVEE/audio EMODB, respectively. In both source tasks and destination tasks of these sets of experiments, our proposed systems are trained for 200 generations. At the beginning of each task, a population of 20 of pathways are randomly generated and are then trained on the source data. In each generation, two paths are randomly selected to train on the source data and then on the destination data for validation. To evaluate one pathway, a pathway is trained with stochastic gradient descent with learning rate 0.02, a mini-batch size of 64 and is trained for $\small{T}$ epochs ($\small{T}$ equals the number of samples counted in training set divided by mini-batch size of 64). The fitness of such pathway is the rate of correct samples on the training set during that period of training time. When completing the calculation of the fitness of two pathways, the pathway with the smaller fitness is replaced by the pathway with the greater one that is then mutated  with equal probability 1/(4$\times$3) per
each candidate of the genotype, a new random integer from range [-2, 2] is added to the current value of the lose candidate. Therefore, the network between tasks is modified as follows: the parameters presented in the best fit pathway, which is evolved on the source data, are fixed and the rest of parameters are randomly reinitialized and a new population of 20 of genotypes are randomly initialized and then are trained/evolved further on the destination data.

\subsection{Experimental Results}

In this section, we report, analyze, and compare experimental results of our proposed system on aforementioned sets of experiments. In all the experiments, we have taken meticulous care to ensure that the test data is never used for training.




\begin{table}[!t]

\begin{center}
\caption{Results of our proposed system when transferring the emotional knowledge from visual eNTERFACE to visual SAVEE in comparison with the best baseline system.}
\label{Resutls of video SAV}
\begin{tabular}{l||cc}
\hline
\textbf{Method} &    \textbf{WAR (\%)}\\
\hline\hline

Fine-Tuned AlexNet \citep{8085174}& 0.85\\
\hline

V$_{\text{SAV}}$&0.89\\
\hline
\hline
\textbf{V$_{\text{eNTER}\rightarrow \text{SAV}}$}&\textbf{0.94}\\
\hline
\end{tabular}
\end{center}
\end{table}

\begin{table}[!t]

\begin{center}
\caption{Results of our proposed system when transferring the emotional knowledge from visual SAVEE to visual eNTERFACE in comparison with the best baseline system.}
\label{Resutls of video eNTER}
\begin{tabular}{l||cc}
\hline
\textbf{Method} &    \textbf{WAR (\%)}\\
\hline\hline

Fine-Tuned Alexnet \citep{8085174}& 0.88\\
\hline


V$_{\text{eNTER}}$&0.88\\
\hline
\hline
\textbf{V$_{\text{SAV}\rightarrow \text{eNTER}}$}&\textbf{0.94}\\
\hline
\end{tabular}
\end{center}
\end{table}

\begin{figure}[!t]
\centering
\subfloat[Confusion matrix of our proposed system when transferring the emotional knowledge from visual eNTERFACE to visual SAVEE]{\includegraphics[width=0.47\linewidth]{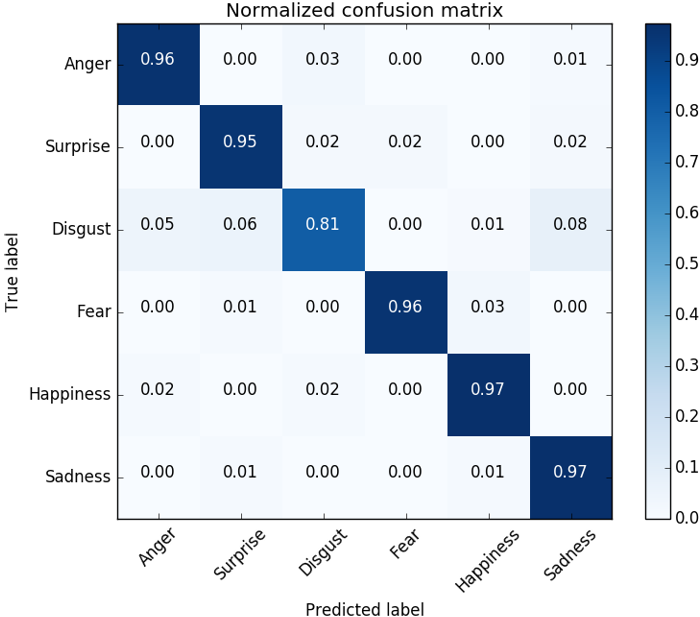}
\label{V_eNTER_SAV_Fitted.png}}
\hfil
\subfloat[Confusion matrix of our proposed system when transferring the emotional knowledge from visual SAVEE to visual eNTERFACE]{\includegraphics[width=0.47\linewidth]{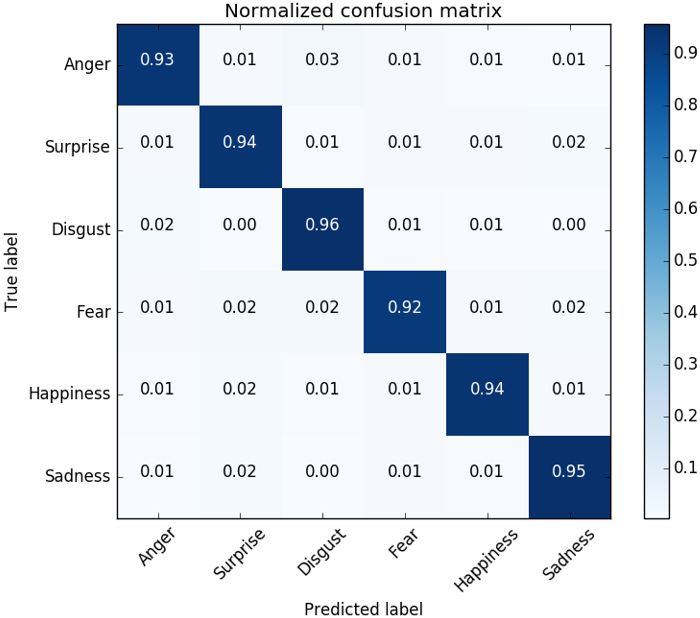}
\label{V_SAV_eNTER_Fitted.png}}
\caption{Illustrates the confusion matrix of our proposed system evaluated on visual SAVEE and visual eNTERFACE}
\label{Visual_Confusion_matrix}
\end{figure}

\label{within-corpus}

In the first two sets of experiments, we conduct experiments to show the robustness of our proposed system when performing under the condition of insufficient data for facial expression recognition. Our proposed system is initially trained on visual eNTERFACE, and is then evolved on visual SAVEE and vice versa. Since SAVEE dataset consists of an additional type of emotion (neutral) compared to eNTERFACE dataset, hence to be consistent between these two datasets, only shared types of emotion including anger, surprise, disgust, fear, happiness, and sadness are detected via these two sets of experiments. The experimental results of our proposed system on this evaluation are shown in the first two rows of Table \ref{cross-corpus experimental results}.

\begin{figure}[!t]
\centering
\subfloat[Testing learning curves on visual SAVEE]{\includegraphics[width=0.49\linewidth]{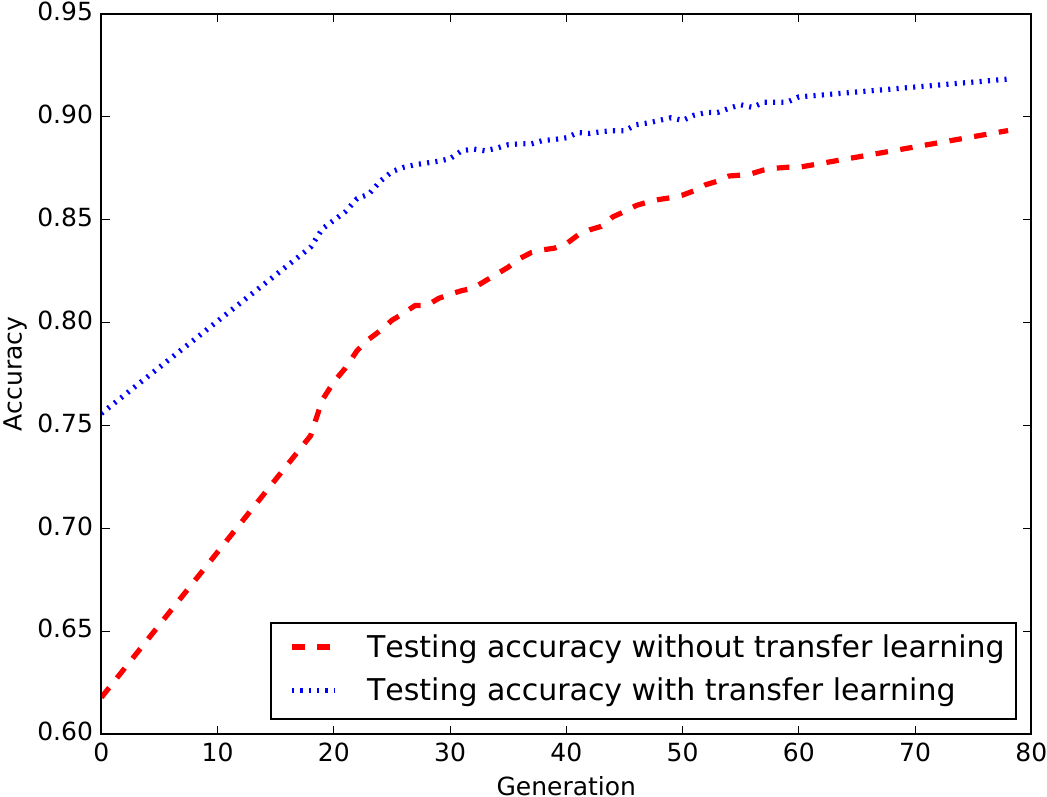}
\label{Learning_curve_V_eNTER_SAV.pdf}}
\hfil
\subfloat[Testing learning curves on visual eNTERFACE]{\includegraphics[width=0.483\linewidth]{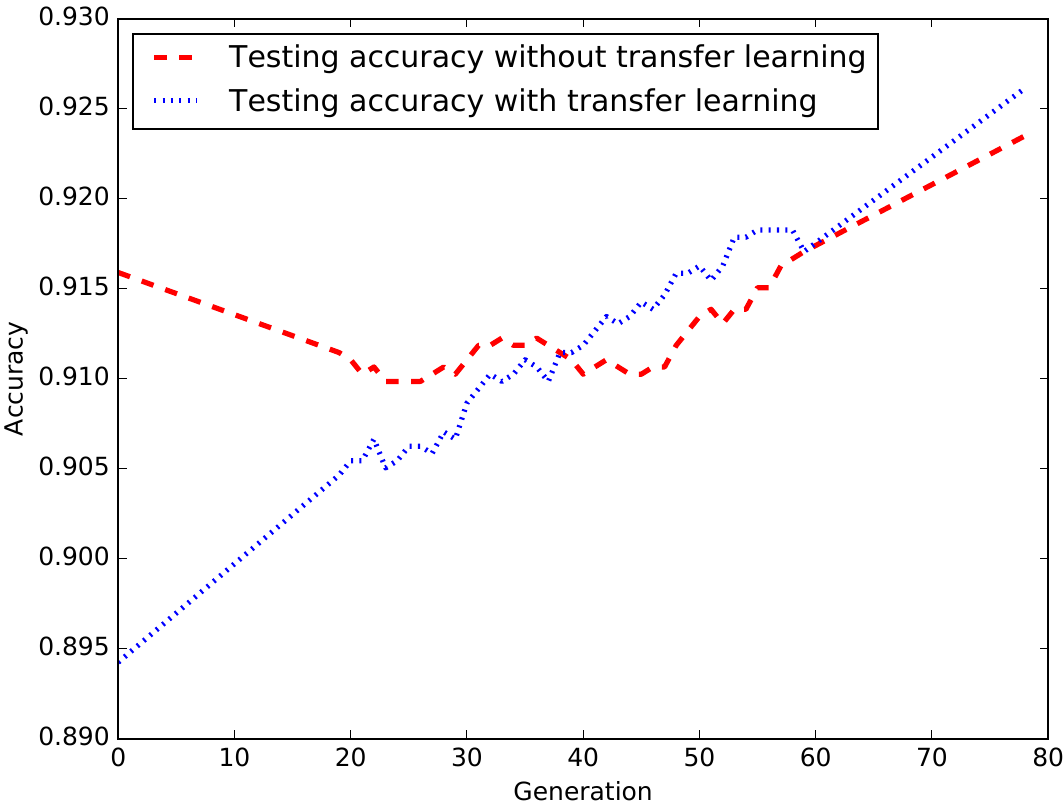}
\label{Learning_curve_V_SAV_eNTER.pdf}}

\caption{Illustrates testing learning curves of our proposed system during learning on visual SAVEE and visual eNTERFACE with and without transfer learning.}  
\label{visual learning curves}
\end{figure}

\begin{figure}[!t]
\centering
\subfloat[Receiver operating characteristic (ROC) curves of individual emotions when transferring the emotional knowledge from visual eNTERFACE to visual SAVEE]{\includegraphics[width=0.48\linewidth]{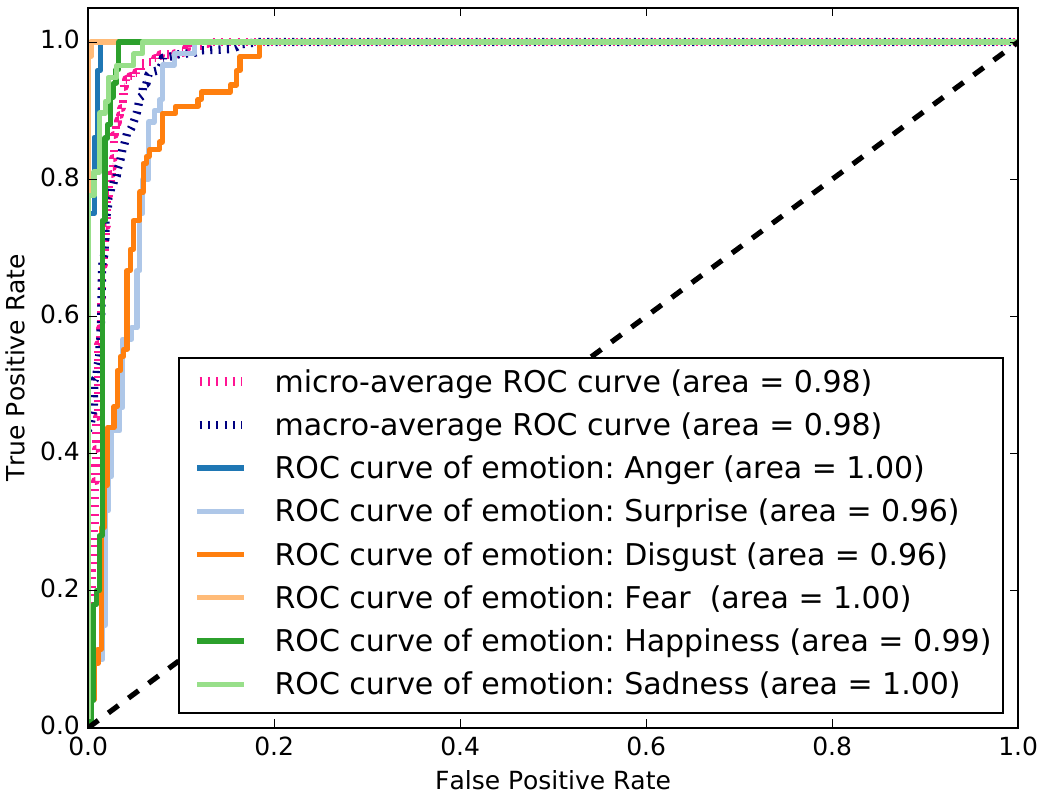}\label{ROC_V_eNTER_SAV_Fitted.png}}
\hfil
\subfloat[Receiver operating characteristic (ROC) curves of individual emotions when transferring the emotional knowledge from visual SAVEE to visual eNTERFACE]{\includegraphics[width=0.48\linewidth]{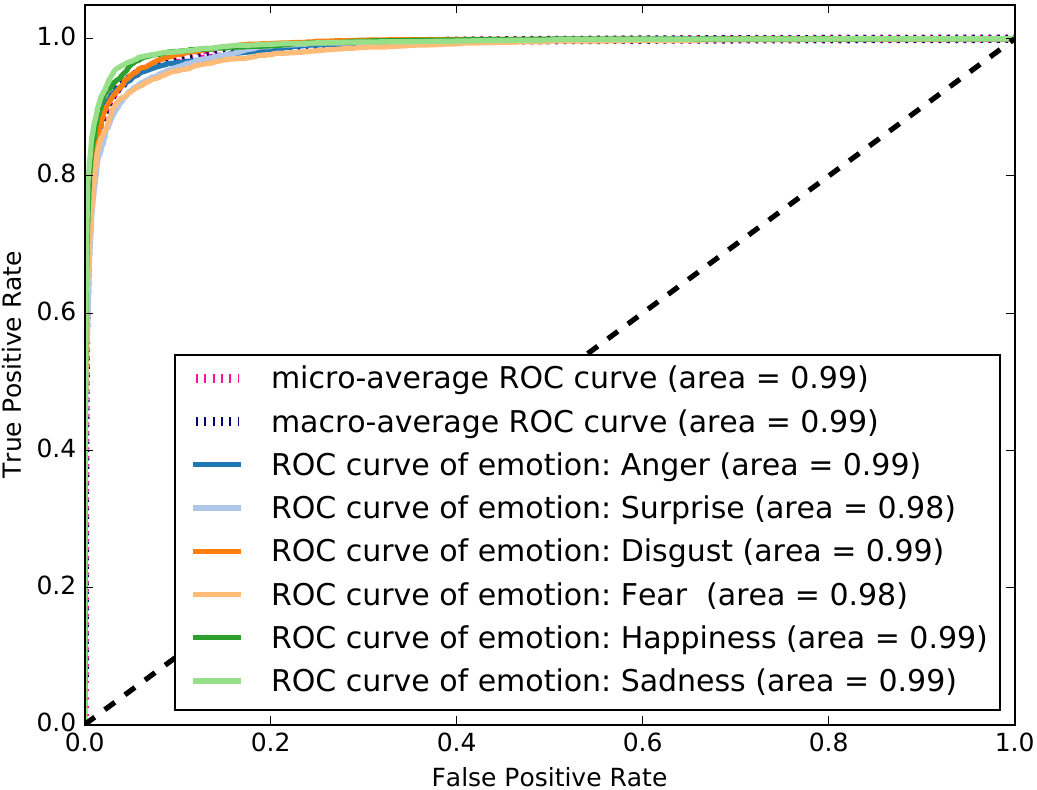}
\label{ROC_V_SAV_eNTER_Fitted.png}}

\caption{Receiver operating characteristic (ROC) curves of our proposed system on visual SAVEE and visual eNTERFACE}
\label{Visual_ROC}
\end{figure}

\textbf{Visual eNTERFACE $\rightarrow$ Visual SAVEE:} As illustrated in Table \ref{cross-corpus experimental results}, our proposed system (V$_{\text{eNTER}\rightarrow \text{SAV}}$), which transfers the emotional knowledge from visual eNTERFACE to visual SAVEE, achieves 94\% of facial expression recognition accuracy in regard to WAR. This accuracy is 5\% and 9\% significant higher than those achieved by our facial expression recognition baseline systems: V$_{\text{SAV}}$ which is trained on visual SAVEE from scratch and the system fine-tuning on AlexNet proposed by \citep{8085174}, respectively (see Table \ref{Resutls of video SAV}).

\textbf{Visual SAVEE $\rightarrow$ Visual eNTERFACE:} To further show the efficiency of our proposed system in solving the issue of insufficient facial expression data, we explore transferring the emotional knowledge from visual SAVEE to visual eNTERFACE. The experimental results are illustrated in the second row of Table \ref{cross-corpus experimental results}. In this setting, our proposed system also performs significantly better than our facial expression recognition baseline systems, achieving the best facial expression recognition accuracy regarding to WAR (94\%), which is 6.5\% and 6.35\% better than those obtained by our facial expression recognition baseline systems: V$_{\text{eNTER}}$ which is trained on visual eNTER from scratch and the system fine-tuning on AlexNet proposed by \citep{8085174} (illustrated in Table \ref{Resutls of video eNTER}).

To depict the performance of individual emotion of these two systems, the confusion matrices of their systems (V$_{\text{eNTER}\rightarrow \text{SAV}}$ and V$_{\text{SAV}\rightarrow \text{eNTER}}$) are illustrated in Fig. \ref{Visual_Confusion_matrix} (a), and Fig. \ref{Visual_Confusion_matrix} (b), respectively. We also visualize the robust performances of each emotion of these two systems, and plot their ROC curves which are illustrated in Fig. \ref{Visual_ROC} (a) and Fig. \ref{Visual_ROC} (b), respectively. Moreover, in order to further visualize the effectiveness of our proposed systems , the different testing learning curves are plotted as illustrated in Fig. \ref{visual learning curves}. As demonstrated in  Fig. \ref{visual learning curves}, the performance of both proposed systems with transfer learning surpasses considerably those without transfer learning.  
 
\label{multiple datasets}

\begin{table}[!t]
\begin{center}
\caption{Results of our proposed system when transferring the emotional knowledge from audio eNTERFACE to audio SAVEE in comparison with the best baseline speech emotion recognition system.}

\label{audio_SAVEE_resutls}
\begin{tabular}{l|cc}
\hline
\textbf{Method} & \textbf{WAR}\\
\hline\hline

AlexNet \citep{8085174} &0.69\\
\hline
A$_{\text{SAV}}$&0.81\\
\hline
\textbf{A$_{\text{eNTER}\rightarrow \text{SAV}}$}&0.85\\
\hline

\end{tabular}
\end{center}

\end{table}

In the first two sets of experiments, we have focused on conducting experiments for facial expression recognition. Through experimental results, we have shown that our proposed facial expression recognition system is vastly boosted when transferring the emotional knowledge from one visual emotion domain to another visual emotion domain using PathNet. The performance of our system is also superior to those of the recent state-of-the-art facial expression recognition systems fine-tuning on the off-the-shelf/pre-trained models. To further demonstrate the effectiveness of our proposed system, we carry out extensive experiments for speech emotion recognition task. 

\textbf{Audio eNTERFACE $\rightarrow$ Audio SAVEE:} In this set of experiments, the emotional knowledge is transferred from audio eNTERFACE to audio SAVEE. Experimental results are shown in Table \ref{audio_SAVEE_resutls}. It can be seen that our proposed system ($A_{\text{eNTER}\rightarrow \text{SAV}}$ with the best speech emotion recognition accuracy of 85\%) demonstrates a significant superior performance over the speech emotion recognition system proposed by \citep{8085174} and our baseline system (A$_{\text{SAV}}$ which have been trained on audio SAVEE from scratch using PathNet) by 16\% and 4\% regarding WAR, respectively.

\textbf{(Audio eNTERFACE and Audio SAVEE) $\rightarrow$ Audio EMODB:} In this set of experiment, our proposed system (A$_{\text{eNTER+SAV}\rightarrow \text{EMO}}$ with the best speech emotion recognition accuracy of 97\%) is trained on audio eNTERFACE and audio SAVEE and the learned emotional knowledge (parameters) presented in the best pathway is then transferred and always involved in training stage on audio EMODB. We apply the leave-one-subject-out cross-validation (LOSOCV) for this set of experiment. The reason we investigate these settings for this set of experiments is that in the first three sets of experiments, we have only explored transferring the emotional knowledge from one emotion domains to another emotion domain, to gain more emotional knowledge, in this set of experiment we explore transferring the emotional knowledge from multiple emotion domains to another emotion domain and use LOSOCV to be compatible with our baseline system. The experimental results are illustrated in Table \ref{Resutls of audio EMODB}. It can be clearly shown in Table \ref{Resutls of audio EMODB} that our proposed system (97\%) performs significantly better than the transfer learning approach based on pre-trained/fine-tuning models \citep{8085174} by 14\% and our baseline system (A$_{\text{EMO}}$ which have been trained on audio EMODB from scratch using PathNet) by 8\% in regard to WAR.

\begin{figure}[!t]
\centering 
\subfloat[Receiver operating characteristic (ROC) curves of individual emotions when transferring the emotional knowledge from audio eNTERFACE to audio SAVEE]{\includegraphics[width=0.479\linewidth]{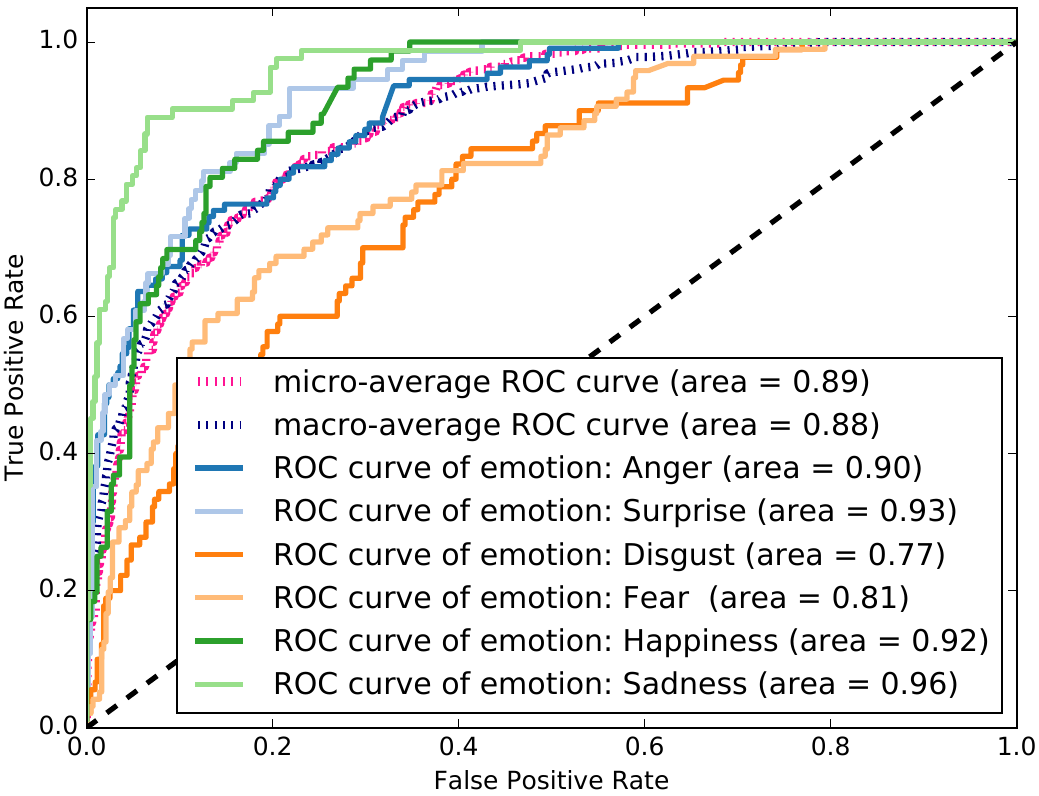}\label{ROC_A_eNTER_SAV_Fitted.png}}
\hfil
\subfloat[Receiver operating characteristic (ROC) curves of individual emotions when transferring the emotional knowledge from audio eNTERFACE and audio SAVEE to audio EMODB]{\includegraphics[width=0.495\linewidth]{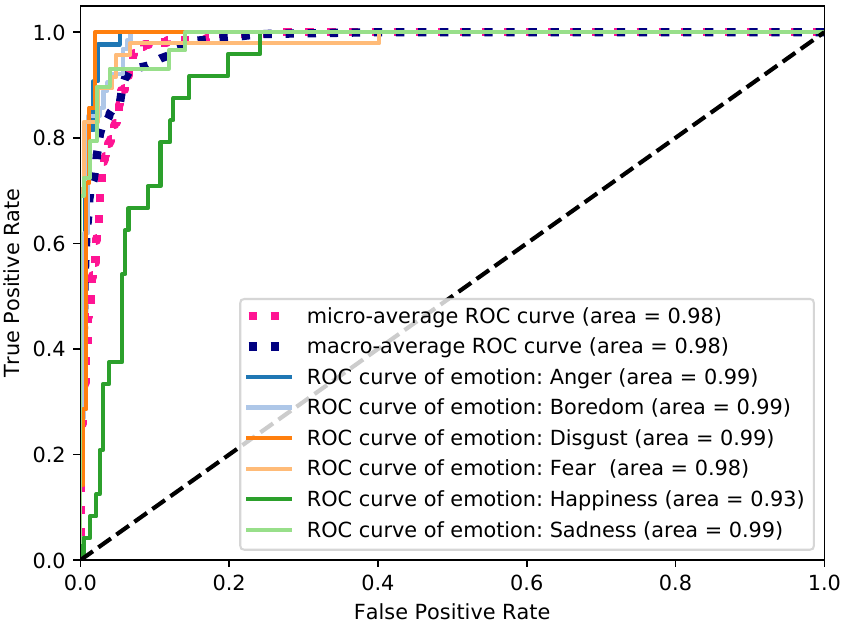}\label{ROC_S_eNTERFACE_SAVEE_T_EMODB__fittetd.pdf}}
\hfil
\caption{We illustrate receiver operating characteristic (ROC) curves of individual emotions evaluated on audio SAVEE and audio EMODB}
\label{ROC_audio}
\end{figure}

\begin{table}[!t]
\begin{center}
\caption{Results of our proposed system when transferring the emotional knowledge from audio eNTERFACE and audio SAVEE to audio EMODB in comparison with the best baseline speech emotion recognition system}
\label{Resutls of audio EMODB}
\begin{tabular}{l|cc}
\hline
\textbf{Method} & \textbf{WAR (\%)}\\
\hline\hline
Fine-Tuned Alexnet-Average \citep{8085174}&0.83\\

\hline
A$_{\text{EMO}}$
&0.89\\
\hline
\textbf{A$_{\text{eNTER+SAV}\rightarrow \text{EMO}}$}&\textbf{0.97}\\
\hline
\end{tabular}
\end{center}

\end{table}

\begin{figure}[!t]
\centering 
\subfloat[Confusion matrix of our proposed system when transferring the emotional knowledge from audio eNTERFACE to audio SAVEE]{\includegraphics[height=0.4\linewidth]{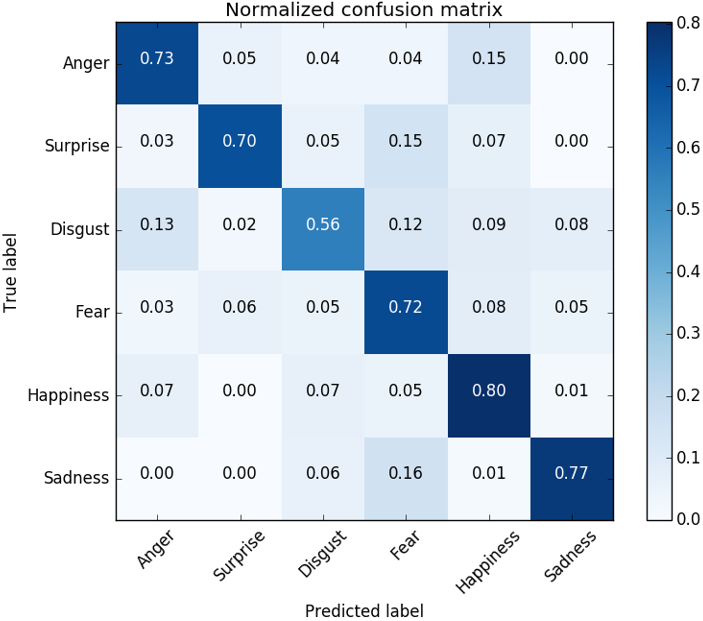}\label{ROC_S_eNTERFACE_SAVEE_T_EMODB__fittetd_.pdf}}
\hfil
\subfloat[Confusion matrix of our proposed system when transferring the emotional knowledge from  audio eNTERFACE and audio SAVEE to audio EMODB]{\includegraphics[height=0.4\linewidth]{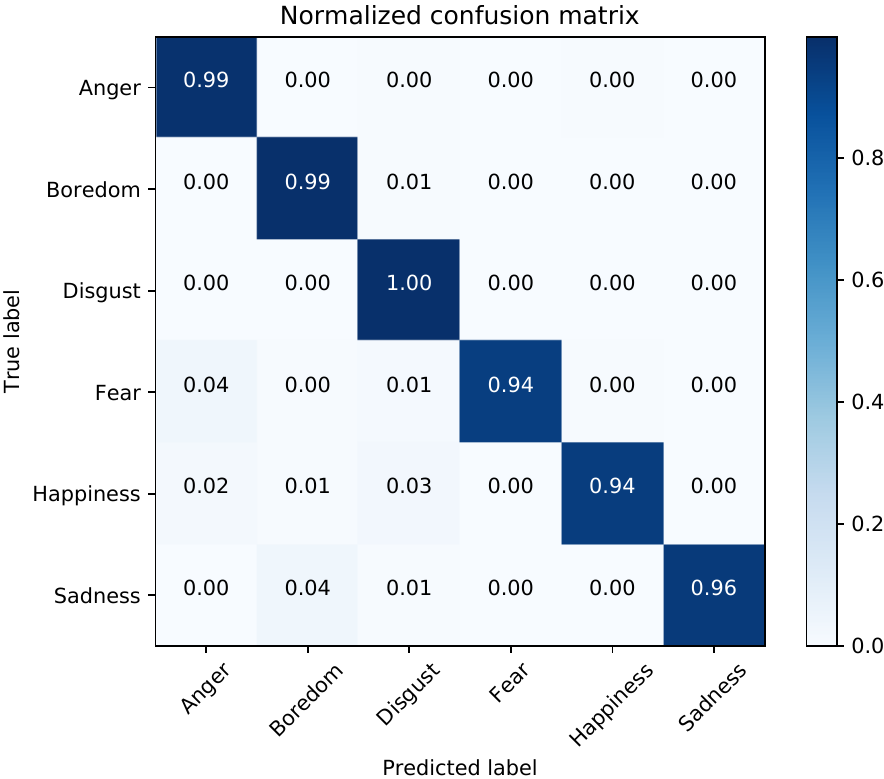}\label{Confusion_matrix_EMODB2.pdf}}

\caption{Illustrates confusion matrix of our proposed system evaluated on audio SAVEE and audio EMODB}
\label{confusion EMODB}
\end{figure}

Similarly, in order to further gain insight into the performances of individual speech emotions of both systems (A$_{\text{eNTER}\rightarrow \text{SAV}}$ and A$_{\text{eNTER+SAV}\rightarrow \text{EMO}}$), we illustrate the confusion matrix of these speech emotion recognition systems (see Fig. \ref{confusion EMODB} (a), and Fig. \ref{confusion EMODB} (b), accordingly). Moreover, we have also plotted their corresponding receiver operating characteristic (ROC) curves as illustrated in Fig. \ref{ROC_audio} (a) and Fig. \ref{ROC_audio} (b). Specifically, our system (A$_{\text{eNTER}\rightarrow \text{SAV}}$) recognizes sadness best with AUC = 0.96, whereas it does not detects disgust and fear so well (with only AUC = 0.77 and AUC = 0.81, accordingly). For other speech emotions including anger, surprise, and happiness our proposed system also demonstrates high-potential performance (with AUC = 0.90, 0.93, and 0.92, respectively) (see Fig. \ref{ROC_audio} (a)). It can also be seen that our A$_{\text{eNTER+SAV}\rightarrow \text{EMO}}$ reveals a great success when detecting speech emotions such as anger, boredom, disgust, fear, and sadness with their corresponding AUC = 0.99, 0.99, 0.99, 0.98, and 0.99, whereas it performs less effectively on the happiness emotion with only AUC = 0.93 (as shown in Fig. \ref{ROC_audio} (b)).

\begin{figure}[!t]
\centering 
{\includegraphics[height=0.47\linewidth]{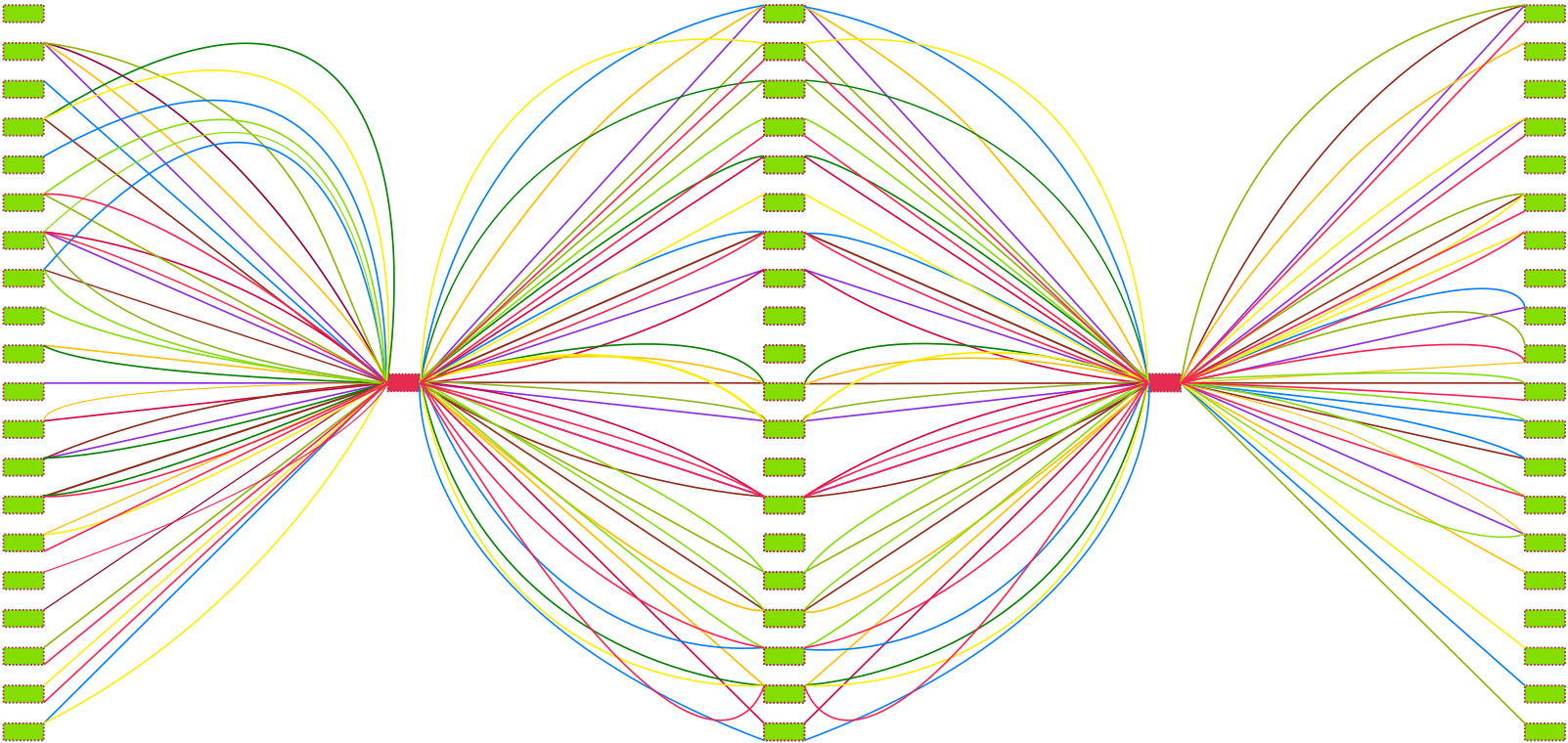}}

\caption{A population of 20 pathways are randomly initialized when learning on the source data (audio eNTERFACE).}
\label{Source_raondom.png}
\end{figure}

\begin{figure}[!t]
\centering 
{\includegraphics[height=0.47\linewidth]{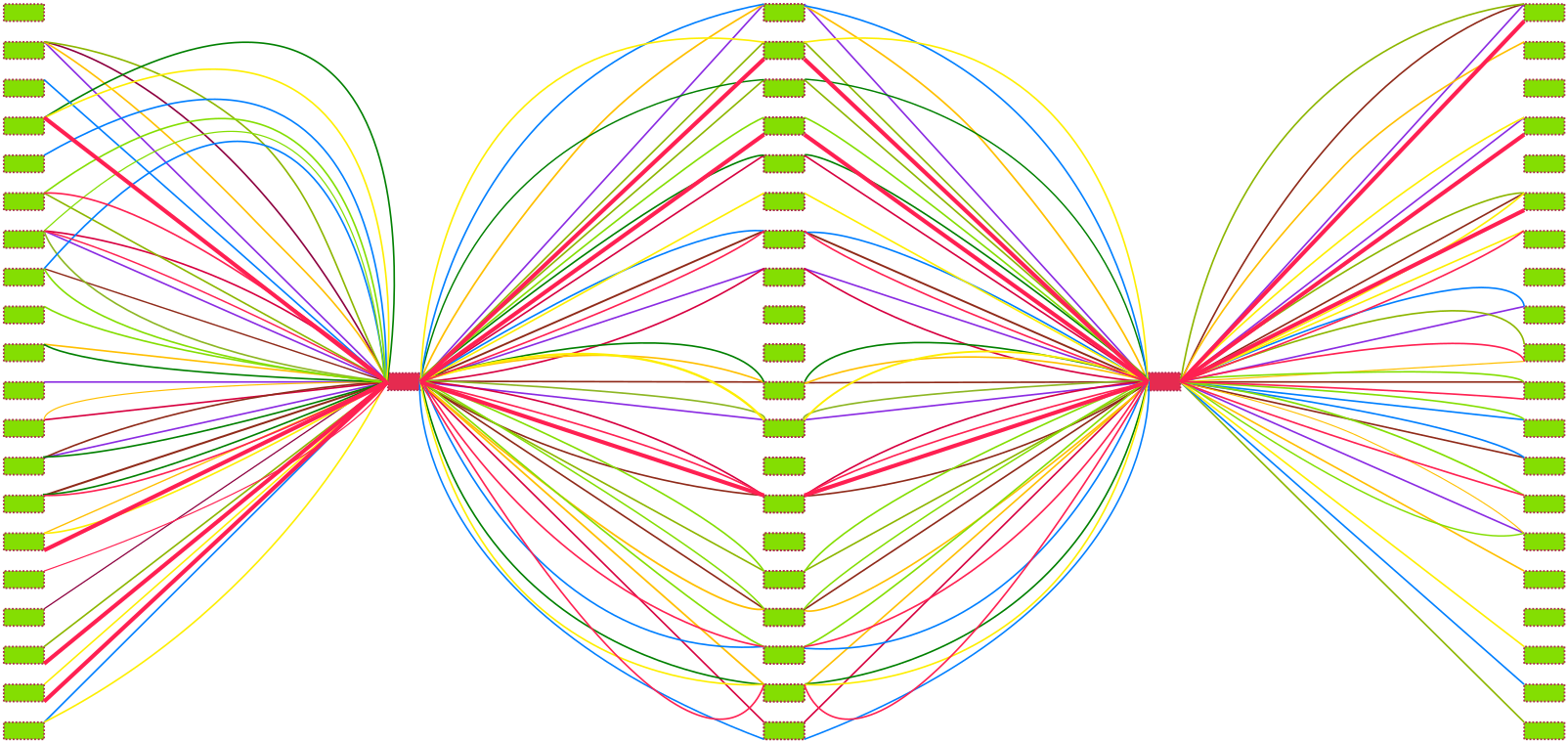}}

\caption{The optimal pathway (highlighted by red colour) is achieved when the training stage is completed on audio eNTERFACE as the source data and the parameters presented in the pathway are fixed.}
\label{Source_best_path_way.png}
\end{figure}

\begin{figure}[!t]
\centering 
{\includegraphics[height=0.47\linewidth]{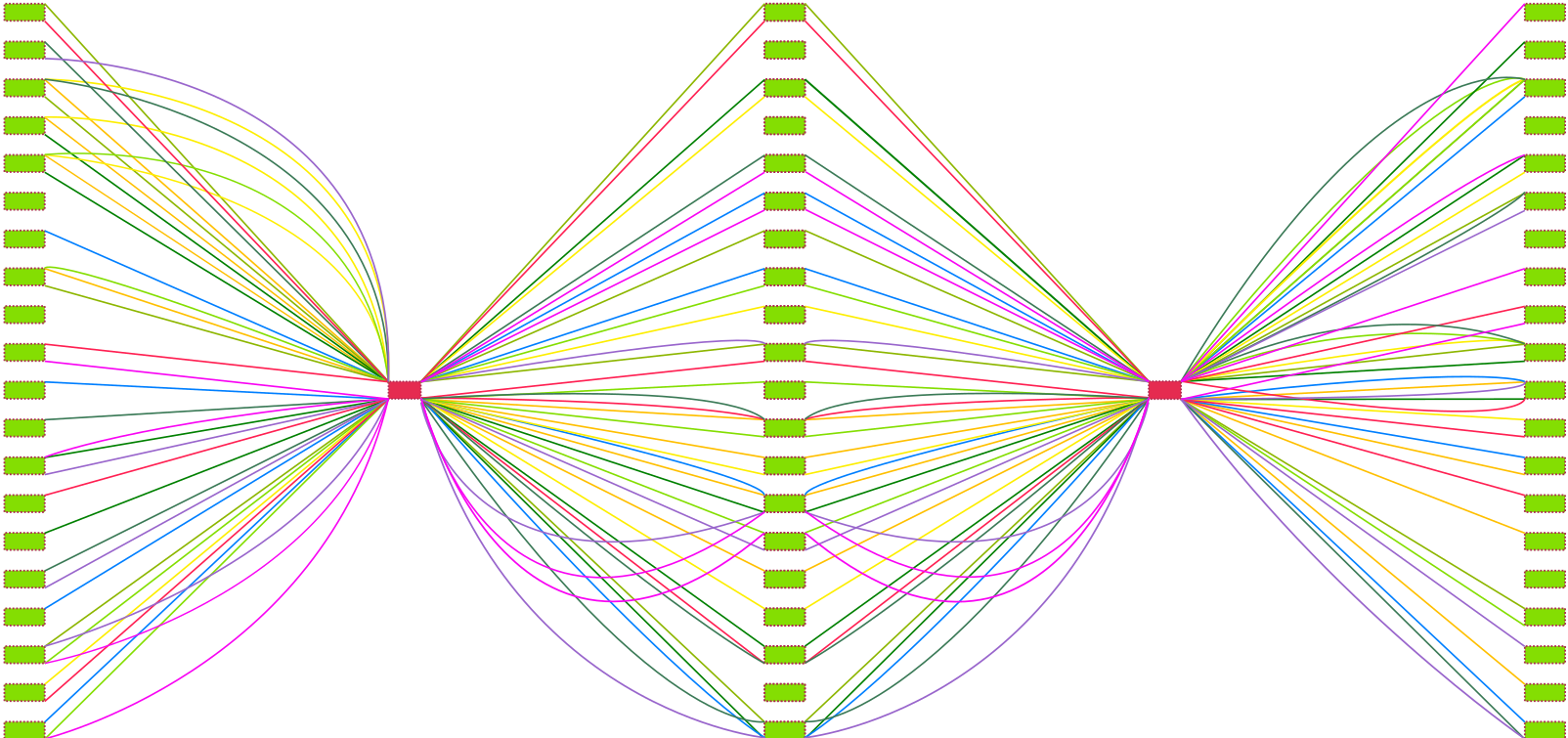}\label{Confusion_matrix_EMODB2__.pdf}}

\caption{A new population of pathways are generated when learning on the destination data (audio SAVEE)}
\label{Destination_random.png}
\end{figure}

\textbf{Discussion:} The reason we achieve significantly better performance when transferring the emotional knowledge from one emotion dataset to another emotion dataset using PathNet is that the emotional knowledge presented in the best PathWay achieved when completed training on one emotion dataset (source data) is now reused as apart of initial emotional knowledge and is always involved in training stage on another set of pathways of PathNet on another emotion dataset (destination data). However, when PathNet is trained on the same emotion dataset (destination data) from scratch, the emotional knowledge is randomly initialized and is learned with only one set of pathways on only one emotion dataset (destination data).

To the best of our knowledge, fine-tuning is arguably the most widely exploited method for transfer learning while working with deep learning architectures. It begins with a pre-trained model on the source task and further trains it on the target task. ImageNet pre-trained models are commonly used for fine-tuning. Compared with training from scratch, fine-tuning a pre-trained off-the-shelf CNN on a target dataset can considerably boost performance, whereas lessening the target annotated data requirements. This transfer learning approach is also considered as one of the most recent state-of-the-art methods. However, as pointed out earlier pre-trained models are still limited in their generalization capability and thus lead to poor performance on novel test sets. The main objective of this paper is to propose an alternative transfer learning approach to address such disadvantages. Therefore, we have primarily compared our transfer learning method using PathNet with a state-of-the-art transfer learning method relying on fine-tuning/pre-trained models as reported in Experiments \& Results Section. The transfer learning method that we have compared against my PathNet based approach using the same datasets is fined-tuned AlexNet. Moreover, unfortunately we could not find any other state of the art transfer learning method which uses the same datasets to make a fair comparison. Our proposed system outperforms the state-of-the-art emotion recognition system relying on the transfer learning method which fine-tunes off-the-shelf/pre-trained models. We believe that this is due to the fact that the representational features that are unrelated to emotion are still retained in off-the-shelf/pre-trained models and the extracted features are also vulnerable to identity variations in these approaches, leading to degrading the performance of the emotion recognition system fine-tuning off-the-shelf/pre-trained models on the emotion dataset.

\begin{figure}[!t]
\centering 

{\includegraphics[height=0.47\linewidth]{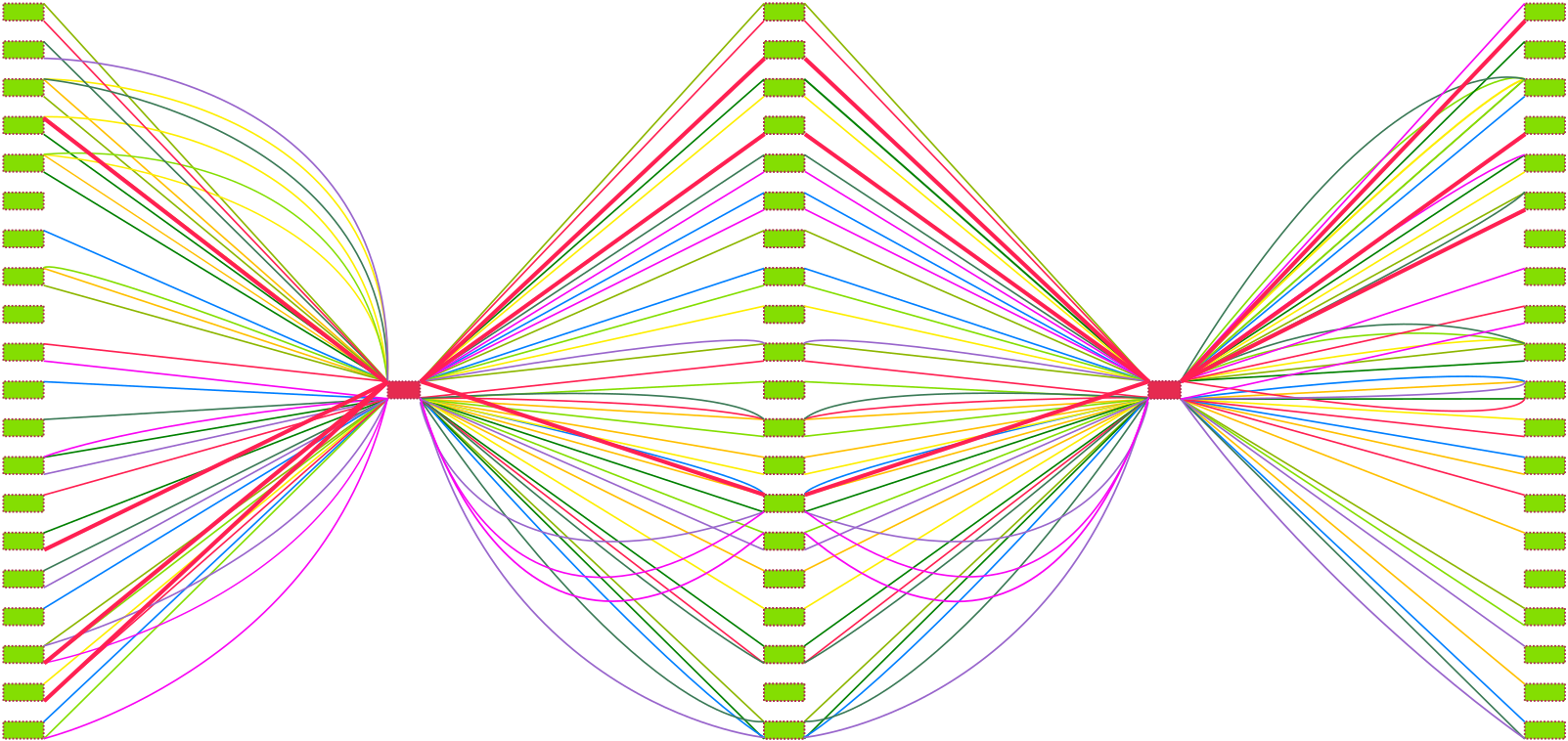}\label{Confusion_matrix_EMODB2_.pdf}}

\caption{The pathway highlighted by red colour, which have been transferred from the model on the source data, are always activated during evolving on the destination task (audio SAVEE), however, their parameters are fixed.}
\label{destination_best_source_pathway_involved.png}
\end{figure}

\begin{figure}[!t]
\centering 
{\includegraphics[height=0.47\linewidth]{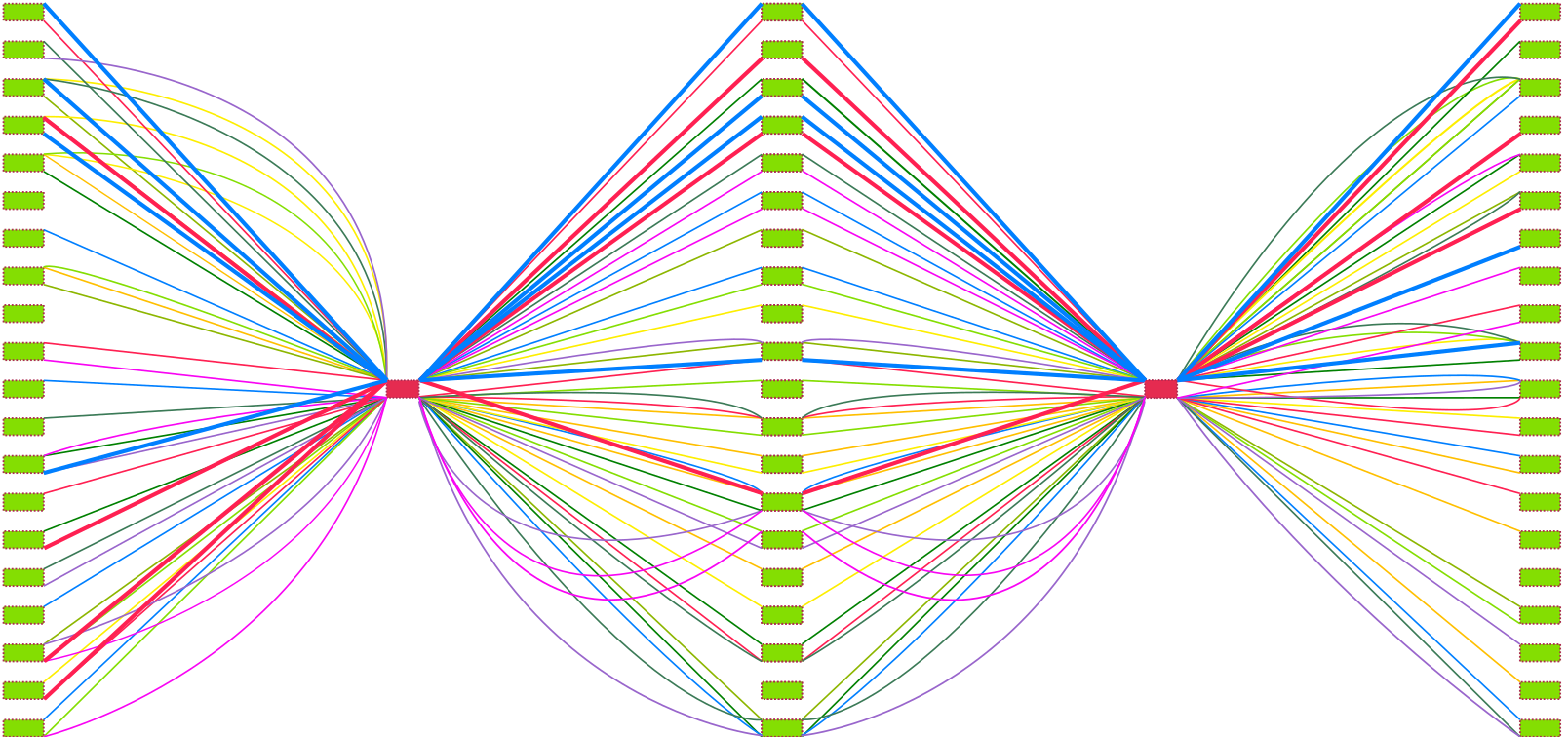}}

\caption{The optimal pathway highlighted by blue colour is achieved when training stage is completed on audio SAVEE (the source data). The paths highlighted by red colour is best pathway gained from the source data.}
\label{destination_best_pathway.png}
\end{figure}

As presented in Proposed Methodology Section, we have designed our PathNet architecture including three layers, each layer consists of 20 random modules. However, only up to 4 modules in each layer are activated during training and 20 random pathways are considered (recall that a pathway is a path that connects activated modules in three layers). It may consider all possible configurations that we can come up with those, we used fully-connected layer for each module. Therefore, there is concern regarding the number of parameters in our model since it appears that this architecture may stack up to a large number of parameters, consequently leading to a possibility that the model is prone to over-fitting. However, as explained clearly in Section 3, in each generation only one pathway is trained and another pathway is subsequently trained. This is because we applied a binary tournament selection genetic algorithm. Therefore, the learnable parameters presented in only two pathways are involved in the training stage using stochastic gradient decent for each generation, but this occurs sequentially. Since up to 4 modules are activated in each layer, 64$\times$64$\times$3 is input for the first layer and 20$\times$1 is input for second and third layer, thus the total number of learnable parameters of one module of one pathway is 64$\times$64$\times$3$\times$20 + 20 for the fist layer, 20$\times$20 + 20 for the second and third layer, resulting in up to (64$\times$64$\times$3$\times$20 + 20 + 20$\times$20 + 20 + 20$\times$20 + 20) $\times$4 $=$ 986,480 parameters for one pathway. While total number of learnable parameters in AlexNet is 60,954,656. Even if we fine-tune only two last fully-connected layers, then the number of learnable parameters is 16,801,783 which is approximately 17 times greater than the number of learnable parameters for one pathway. In the future work, we may consider the use of convolution operation in each module rather than linear multiplication. This can help capture spatial information and significantly reduce the parameters presented in one pathway.

Additionally, as presented in Section 3.3.2 (Pathway Evolution and Transfer Learning Approach) and also discussed in previous paragraph, in each generation two random pathways are selected among a population of pathways to sequentially train on source dataset. A pathway that has better recognition accuracy is kept and mutated. This means that different emotions (e.g., different labels) are simultaneously used in one pathway and we base on overall performance of one pathway to make a decision whether it is removed or mutated (e.g., competed against a new pathway). If different emotions (e.g., six emotions in this work) are associated with different pathways, for instance one pathway is learned on one emotion, we need a set of six pathways for six emotions and calculate six corresponding recognition accuracy, then we need another set of six pathways. A good point by doing this manner is that when we finish training for a number of generations, we can achieve a best set of six pathways and parameters presented in this set of six pathways are reused and transferred to train simultaneously on six emotions of destination data. Due to six different best sets of parameters from different emotions now involved in one training, we may obtain a better performance overall. However, obviously twelve pathways in each generation are considered, while only two are used in our design of training manner.

\textbf{Visualization:} To understand more how to achieve the best pathways when training our proposed system on the source domain and on the destination domain, we specifically visualize step by step how to obtain such pathways when our model is learned on audio eNTERFACE and the learned emotional knowledge is transferred to the model which is further learned on audio SAVEE as shown in Fig. \ref{Source_raondom.png}, Fig. \ref{Source_best_path_way.png}, Fig. \ref{Destination_random.png}, Fig. \ref{destination_best_source_pathway_involved.png}, and Fig. \ref{destination_best_pathway.png}.

\section{Conclusion}
Progress in emotion recognition research has been hindered by the lack of the large amounts of labeled emotion data. To overcome this problem, various studies have widely explored the use of transfer learning approach based on pre-trained/fine-tuning models, however, the proposed approaches have been still suffering from issues such as discarding prior learned information. In this paper, we have proposed utilizing an alternative transfer learning technique using PathNet which is a neural network algorithm that uses agents embedded in the neural network whose task is to discover which parts of the network to reuse for new tasks, leading to successfully addressing the above-mentioned challenges. To verify performance of our proposed architecture, we have conducted various sets of experiments including, transferring the emotional knowledge from one emotion dataset into another, and transferring the learned emotional knowledge from multiple emotion datasets into one another. Experimental results on three datasets: eNTERFACE, SAVEE, and EMO-DB have indicated that our proposed system performs well under the conditions of insufficient emotion data, and significantly better than the recent transfer learning techniques exploiting fine-tuning/pre-trained models.
\section{Acknowledgement}
This research was supported by an Australian Research Council (ARC) Discovery grant DP140100793.

\bibliographystyle{model2-names.bst}\biboptions{authoryear}
\bibliography{refs}

\begin{thebibliography}{41}
\expandafter\ifx\csname natexlab\endcsname\relax\def\natexlab#1{#1}\fi
\providecommand{\url}[1]{\texttt{#1}}
\providecommand{\href}[2]{#2}
\providecommand{\path}[1]{#1}
\providecommand{\DOIprefix}{doi:}
\providecommand{\ArXivprefix}{arXiv:}
\providecommand{\URLprefix}{URL: }
\providecommand{\Pubmedprefix}{pmid:}
\providecommand{\doi}[1]{\href{http://dx.doi.org/#1}{\path{#1}}}
\providecommand{\Pubmed}[1]{\href{pmid:#1}{\path{#1}}}
\providecommand{\bibinfo}[2]{#2}
\ifx\xfnm\relax \def\xfnm[#1]{\unskip,\space#1}\fi
\bibitem[{{Abbasnejad} et~al.(2017){Abbasnejad}, {Sridharan}, {Nguyen},
  {Denman}, {Fookes} and {Lucey}}]{Abbasnejad_2017_ICCV}
\bibinfo{author}{{Abbasnejad}, I.}, \bibinfo{author}{{Sridharan}, S.},
  \bibinfo{author}{{Nguyen}, D.}, \bibinfo{author}{{Denman}, S.},
  \bibinfo{author}{{Fookes}, C.}, \bibinfo{author}{{Lucey}, S.},
  \bibinfo{year}{2017}.
\newblock \bibinfo{title}{Using synthetic data to improve facial expression
  analysis with 3d convolutional networks}, in: \bibinfo{booktitle}{2017 IEEE
  International Conference on Computer Vision Workshops (ICCVW)}, pp.
  \bibinfo{pages}{1609--1618}.
\bibitem[{Badshah et~al.(2017)Badshah, Ahmad, Rahim and Baik}]{7883728}
\bibinfo{author}{Badshah, A.M.}, \bibinfo{author}{Ahmad, J.},
  \bibinfo{author}{Rahim, N.}, \bibinfo{author}{Baik, S.W.},
  \bibinfo{year}{2017}.
\newblock \bibinfo{title}{Speech emotion recognition from spectrograms with
  deep convolutional neural network}, in: \bibinfo{booktitle}{2017
  International Conference on Platform Technology and Service (PlatCon)}, pp.
  \bibinfo{pages}{1--5}.
\bibitem[{Burkhardt et~al.(2005)Burkhardt, Paeschke, Rolfes, Sendlmeier and
  Weiss}]{Burkhardt05adatabase}
\bibinfo{author}{Burkhardt, F.}, \bibinfo{author}{Paeschke, A.},
  \bibinfo{author}{Rolfes, M.}, \bibinfo{author}{Sendlmeier, W.},
  \bibinfo{author}{Weiss, B.}, \bibinfo{year}{2005}.
\newblock \bibinfo{title}{A database of german emotional speech}, in:
  \bibinfo{booktitle}{in Proceedings of Interspeech, Lissabon}, pp.
  \bibinfo{pages}{1517--1520}.
\bibitem[{Chang and Scherer(2017)}]{7952656}
\bibinfo{author}{Chang, J.}, \bibinfo{author}{Scherer, S.},
  \bibinfo{year}{2017}.
\newblock \bibinfo{title}{Learning representations of emotional speech with
  deep convolutional generative adversarial networks}, in:
  \bibinfo{booktitle}{2017 ICASSP}, pp. \bibinfo{pages}{2746--2750}.
\bibitem[{Deng et~al.(2017a)Deng, Frühholz, Zhang and Schuller}]{7879177}
\bibinfo{author}{Deng, J.}, \bibinfo{author}{Frühholz, S.},
  \bibinfo{author}{Zhang, Z.}, \bibinfo{author}{Schuller, B.},
  \bibinfo{year}{2017}a.
\newblock \bibinfo{title}{Recognizing emotions from whispered speech based on
  acoustic feature transfer learning}.
\newblock \bibinfo{journal}{IEEE Access} \bibinfo{volume}{5},
  \bibinfo{pages}{5235--5246}.
\bibitem[{Deng et~al.(2017b)Deng, Xu, Zhang, Frühholz and Schuller}]{7862157}
\bibinfo{author}{Deng, J.}, \bibinfo{author}{Xu, X.}, \bibinfo{author}{Zhang,
  Z.}, \bibinfo{author}{Frühholz, S.}, \bibinfo{author}{Schuller, B.},
  \bibinfo{year}{2017}b.
\newblock \bibinfo{title}{Universum autoencoder-based domain adaptation for
  speech emotion recognition}.
\newblock \bibinfo{journal}{IEEE Signal Processing Letters}
  \bibinfo{volume}{24}, \bibinfo{pages}{500--504}.
\bibitem[{Eyben(2016)}]{eyben2015real}
\bibinfo{author}{Eyben, F.}, \bibinfo{year}{2016}.
\newblock \bibinfo{title}{Real-time speech and music classification by large
  audio feature space extraction}.
\newblock \bibinfo{publisher}{Springer}, \bibinfo{address}{New York, NY, USA}.
\bibitem[{Eyben et~al.(2016)Eyben, Scherer, Schuller, Sundberg, André, Busso,
  Devillers, Epps, Laukka, Narayanan and Truong}]{7160715}
\bibinfo{author}{Eyben, F.}, \bibinfo{author}{Scherer, K.R.},
  \bibinfo{author}{Schuller, B.W.}, \bibinfo{author}{Sundberg, J.},
  \bibinfo{author}{André, E.}, \bibinfo{author}{Busso, C.},
  \bibinfo{author}{Devillers, L.Y.}, \bibinfo{author}{Epps, J.},
  \bibinfo{author}{Laukka, P.}, \bibinfo{author}{Narayanan, S.S.},
  \bibinfo{author}{Truong, K.P.}, \bibinfo{year}{2016}.
\newblock \bibinfo{title}{The geneva minimalistic acoustic parameter set
  (gemaps) for voice research and affective computing}.
\newblock \bibinfo{journal}{IEEE Transactions on Affective Computing}
  \bibinfo{volume}{7}, \bibinfo{pages}{190--202}.
\bibitem[{Fan et~al.(2016)Fan, Lu, Li and Liu}]{Fan:2016:VER:2993148.2997632}
\bibinfo{author}{Fan, Y.}, \bibinfo{author}{Lu, X.}, \bibinfo{author}{Li, D.},
  \bibinfo{author}{Liu, Y.}, \bibinfo{year}{2016}.
\newblock \bibinfo{title}{Video-based emotion recognition using {CNN-RNN} and
  {C3D} hybrid networks}, in: \bibinfo{booktitle}{Proceedings of the 18th ACM
  International Conference on Multimodal Interaction},
  \bibinfo{publisher}{ACM}, \bibinfo{address}{New York, NY, USA}. pp.
  \bibinfo{pages}{445--450}.
\bibitem[{Fernando et~al.(2017)Fernando, Banarse, Blundell, Zwols, Ha, Rusu,
  Pritzel and Wierstra}]{DBLP:journals/corr/FernandoBBZHRPW17}
\bibinfo{author}{Fernando, C.}, \bibinfo{author}{Banarse, D.},
  \bibinfo{author}{Blundell, C.}, \bibinfo{author}{Zwols, Y.},
  \bibinfo{author}{Ha, D.}, \bibinfo{author}{Rusu, A.A.},
  \bibinfo{author}{Pritzel, A.}, \bibinfo{author}{Wierstra, D.},
  \bibinfo{year}{2017}.
\newblock \bibinfo{title}{Pathnet: Evolution channels gradient descent in super
  neural networks}.
\newblock \bibinfo{journal}{CoRR} \bibinfo{volume}{abs/1701.08734}.
\bibitem[{Gideon et~al.(2017)Gideon, Khorram, Aldeneh, Dimitriadis and
  Provost}]{DBLP:journals/corr/GideonKADP17}
\bibinfo{author}{Gideon, J.}, \bibinfo{author}{Khorram, S.},
  \bibinfo{author}{Aldeneh, Z.}, \bibinfo{author}{Dimitriadis, D.},
  \bibinfo{author}{Provost, E.M.}, \bibinfo{year}{2017}.
\newblock \bibinfo{title}{Progressive neural networks for transfer learning in
  emotion recognition}.
\newblock \bibinfo{journal}{CoRR} \bibinfo{volume}{abs/1706.03256}.
\bibitem[{Haq and Jackson(2010)}]{HaqJackson_MachineAudition10}
\bibinfo{author}{Haq, S.}, \bibinfo{author}{Jackson, P.}, \bibinfo{year}{2010}.
\newblock \bibinfo{title}{Machine Audition: Principles, Algorithms and
  Systems}. \bibinfo{publisher}{IGI Global}, \bibinfo{address}{Hershey PA}.
  chapter \bibinfo{chapter}{Multimodal Emotion Recognition}.
\newblock pp. \bibinfo{pages}{398--423}.
\bibitem[{Harvey(2011)}]{Harvey:2009:MGA:2017762.2017781}
\bibinfo{author}{Harvey, I.}, \bibinfo{year}{2011}.
\newblock \bibinfo{title}{The microbial genetic algorithm}, in:
  \bibinfo{booktitle}{2009 ECRL}, \bibinfo{publisher}{Springer-Verlag},
  \bibinfo{address}{Berlin, Heidelberg}. pp. \bibinfo{pages}{126--133}.
\bibitem[{Hasani and Mahoor(2017)}]{8015016}
\bibinfo{author}{Hasani, B.}, \bibinfo{author}{Mahoor, M.H.},
  \bibinfo{year}{2017}.
\newblock \bibinfo{title}{Facial expression recognition using enhanced deep
  3{D} convolutional neural networks}, in: \bibinfo{booktitle}{2017 CVPRW}, pp.
  \bibinfo{pages}{2278--2288}.
\bibitem[{Hermansky et~al.(1992)Hermansky, Morgan, Bayya and
  Kohn}]{Hermansky:1992:RSA:1895550.1895585}
\bibinfo{author}{Hermansky, H.}, \bibinfo{author}{Morgan, N.},
  \bibinfo{author}{Bayya, A.}, \bibinfo{author}{Kohn, P.},
  \bibinfo{year}{1992}.
\newblock \bibinfo{title}{Rasta-plp speech analysis technique}, in:
  \bibinfo{booktitle}{1992 ICASSP}, \bibinfo{publisher}{IEEE Computer Society},
  \bibinfo{address}{Washington, DC, USA}. pp. \bibinfo{pages}{121--124}.
\bibitem[{Kaya et~al.(2017)Kaya, Gürpınar and Salah}]{KAYA201766}
\bibinfo{author}{Kaya, H.}, \bibinfo{author}{Gürpınar, F.},
  \bibinfo{author}{Salah, A.A.}, \bibinfo{year}{2017}.
\newblock \bibinfo{title}{Video-based emotion recognition in the wild using
  deep transfer learning and score fusion}.
\newblock \bibinfo{journal}{Image and Vision Computing} \bibinfo{volume}{65},
  \bibinfo{pages}{66 -- 75}.
\newblock \bibinfo{note}{Multimodal Sentiment Analysis and Mining in the Wild
  Image and Vision Computing}.
\bibitem[{Kim et~al.(2017a)Kim, Englebienne, Truong and Evers}]{kim2017acmmm}
\bibinfo{author}{Kim, J.}, \bibinfo{author}{Englebienne, G.},
  \bibinfo{author}{Truong, K.P.}, \bibinfo{author}{Evers, V.},
  \bibinfo{year}{2017}a.
\newblock \bibinfo{title}{Deep temporal models using identity skip-connections
  for speech emotion recognition}, in: \bibinfo{booktitle}{Proceedings of ACM
  Multimedia}, pp. \bibinfo{pages}{1006--1013}.
\bibitem[{Kim et~al.(2017b)Kim, Englebienne, Truong and
  Evers}]{kim2017interspeech}
\bibinfo{author}{Kim, J.}, \bibinfo{author}{Englebienne, G.},
  \bibinfo{author}{Truong, K.P.}, \bibinfo{author}{Evers, V.},
  \bibinfo{year}{2017}b.
\newblock \bibinfo{title}{Towards speech emotion recognition 'in the wild'
  using aggregated corpora and deep multi-task learning}, in:
  \bibinfo{booktitle}{Proceedings of the INTERSPEECH}, pp.
  \bibinfo{pages}{1113--1117}.
\bibitem[{Kim et~al.(2017c)Kim, Truong, Englebienne and Evers}]{kim2017acii}
\bibinfo{author}{Kim, J.}, \bibinfo{author}{Truong, K.},
  \bibinfo{author}{Englebienne, G.}, \bibinfo{author}{Evers, V.},
  \bibinfo{year}{2017}c.
\newblock \bibinfo{title}{Learning spectro-temporal features with 3{DCNN}s for
  speech emotion recognition}, in: \bibinfo{booktitle}{Proceedings of
  International Conference on Affective Computing and Intelligent Interaction},
  pp. \bibinfo{pages}{383--388}.
\bibitem[{Krizhevsky and Hinton(2009)}]{Krizhevsky09}
\bibinfo{author}{Krizhevsky, A.}, \bibinfo{author}{Hinton, G.},
  \bibinfo{year}{2009}.
\newblock \bibinfo{title}{Learning multiple layers of features from tiny
  images}.
\newblock \bibinfo{journal}{Master's thesis, Department of Computer Science,
  University of Toronto} .
\bibitem[{Latif et~al.(2018)Latif, Rana, Younis, Qadir and
  Epps}]{DBLP:journals/corr/abs-1801-06353}
\bibinfo{author}{Latif, S.}, \bibinfo{author}{Rana, R.},
  \bibinfo{author}{Younis, S.}, \bibinfo{author}{Qadir, J.},
  \bibinfo{author}{Epps, J.}, \bibinfo{year}{2018}.
\newblock \bibinfo{title}{Cross corpus speech emotion classification- an
  effective transfer learning technique}.
\newblock \bibinfo{journal}{CoRR} \bibinfo{volume}{abs/1801.06353}.
\bibitem[{LeCun et~al.(2015)LeCun, Bengio and Hinton}]{Denet2015}
\bibinfo{author}{LeCun, Y.}, \bibinfo{author}{Bengio, Y.},
  \bibinfo{author}{Hinton, G.}, \bibinfo{year}{2015}.
\newblock \bibinfo{title}{Deep learning}.
\newblock \bibinfo{journal}{NATURE} \bibinfo{volume}{521},
  \bibinfo{pages}{436--444}.
\bibitem[{Lee et~al.(2017)Lee, Kim, Jun, Ha and Zhang}]{NIPS2017_7051}
\bibinfo{author}{Lee, S.W.}, \bibinfo{author}{Kim, J.H.}, \bibinfo{author}{Jun,
  J.}, \bibinfo{author}{Ha, J.W.}, \bibinfo{author}{Zhang, B.T.},
  \bibinfo{year}{2017}.
\newblock \bibinfo{title}{Overcoming catastrophic forgetting by incremental
  moment matching}, in: \bibinfo{editor}{Guyon, I.}, \bibinfo{editor}{Luxburg,
  U.V.}, \bibinfo{editor}{Bengio, S.}, \bibinfo{editor}{Wallach, H.},
  \bibinfo{editor}{Fergus, R.}, \bibinfo{editor}{Vishwanathan, S.},
  \bibinfo{editor}{Garnett, R.} (Eds.), \bibinfo{booktitle}{2017 NIPS}.
  \bibinfo{publisher}{Curran Associates, Inc.}, pp.
  \bibinfo{pages}{4655--4665}.
\bibitem[{Martin et~al.(2006)Martin, Kotsia, Macq and Pitas}]{1623803}
\bibinfo{author}{Martin, O.}, \bibinfo{author}{Kotsia, I.},
  \bibinfo{author}{Macq, B.}, \bibinfo{author}{Pitas, I.},
  \bibinfo{year}{2006}.
\newblock \bibinfo{title}{The e{NTERFACE}' 05 audio-visual emotion database},
  in: \bibinfo{booktitle}{Data Engineering Workshops, 2006. Proceedings. 22nd
  International Conference on}, pp. \bibinfo{pages}{8--8}.
\bibitem[{Netzer et~al.(2011)Netzer, Wang, Coates, Bissacco, Wu and Ng}]{37648}
\bibinfo{author}{Netzer, Y.}, \bibinfo{author}{Wang, T.},
  \bibinfo{author}{Coates, A.}, \bibinfo{author}{Bissacco, A.},
  \bibinfo{author}{Wu, B.}, \bibinfo{author}{Ng, A.Y.}, \bibinfo{year}{2011}.
\newblock \bibinfo{title}{Reading digits in natural images with unsupervised
  feature learning}, in: \bibinfo{booktitle}{NIPS Workshop on Deep Learning and
  Unsupervised Feature Learning 2011}, pp. \bibinfo{pages}{1--3}.
\bibitem[{Ng et~al.(2015)Ng, Nguyen, Vonikakis and
  Winkler}]{Ng:2015:DLE:2818346.2830593}
\bibinfo{author}{Ng, H.W.}, \bibinfo{author}{Nguyen, V.D.},
  \bibinfo{author}{Vonikakis, V.}, \bibinfo{author}{Winkler, S.},
  \bibinfo{year}{2015}.
\newblock \bibinfo{title}{Deep learning for emotion recognition on small
  datasets using transfer learning}, in: \bibinfo{booktitle}{ICMI},
  \bibinfo{publisher}{ACM}, \bibinfo{address}{New York, NY, USA}. pp.
  \bibinfo{pages}{443--449}.
\bibitem[{{Nguyen} et~al.(2018){Nguyen}, {Nguyen}, {Sridharan}, {Abbasnejad},
  {Dean} and {Fookes}}]{8545411}
\bibinfo{author}{{Nguyen}, D.}, \bibinfo{author}{{Nguyen}, K.},
  \bibinfo{author}{{Sridharan}, S.}, \bibinfo{author}{{Abbasnejad}, I.},
  \bibinfo{author}{{Dean}, D.}, \bibinfo{author}{{Fookes}, C.},
  \bibinfo{year}{2018}.
\newblock \bibinfo{title}{Meta transfer learning for facial emotion
  recognition}, in: \bibinfo{booktitle}{ICPR}, pp. \bibinfo{pages}{3543--3548}.
\bibitem[{Nguyen et~al.(2018)Nguyen, Nguyen, Sridharan, Dean and
  Fookes}]{NGUYEN2018}
\bibinfo{author}{Nguyen, D.}, \bibinfo{author}{Nguyen, K.},
  \bibinfo{author}{Sridharan, S.}, \bibinfo{author}{Dean, D.},
  \bibinfo{author}{Fookes, C.}, \bibinfo{year}{2018}.
\newblock \bibinfo{title}{Deep spatio-temporal feature fusion with compact
  bilinear pooling for multimodal emotion recognition}.
\newblock \bibinfo{journal}{Computer Vision and Image Understanding} .
\bibitem[{Nguyen et~al.(2017)Nguyen, Nguyen, Sridharan, Ghasemi, Dean and
  Fookes}]{7926723}
\bibinfo{author}{Nguyen, D.}, \bibinfo{author}{Nguyen, K.},
  \bibinfo{author}{Sridharan, S.}, \bibinfo{author}{Ghasemi, A.},
  \bibinfo{author}{Dean, D.}, \bibinfo{author}{Fookes, C.},
  \bibinfo{year}{2017}.
\newblock \bibinfo{title}{Deep spatio-temporal features for multimodal emotion
  recognition}, in: \bibinfo{booktitle}{2017 IEEE Winter Conference on
  Applications of Computer Vision (WACV)}, pp. \bibinfo{pages}{1215--1223}.
\bibitem[{Nguyen et~al.(2020)Nguyen, Sridharan, Nguyen, Denman, Tran, Zeng and
  Fookes}]{nguyen2020joint}
\bibinfo{author}{Nguyen, D.}, \bibinfo{author}{Sridharan, S.},
  \bibinfo{author}{Nguyen, D.T.}, \bibinfo{author}{Denman, S.},
  \bibinfo{author}{Tran, S.N.}, \bibinfo{author}{Zeng, R.},
  \bibinfo{author}{Fookes, C.}, \bibinfo{year}{2020}.
\newblock \bibinfo{title}{Joint deep cross-domain transfer learning for emotion
  recognition}.
\newblock \href{http://arxiv.org/abs/2003.11136}{\tt arXiv:2003.11136}.
\bibitem[{Pao et~al.(2006)Pao, Chen, Yeh and Li}]{1699080}
\bibinfo{author}{Pao, T.L.}, \bibinfo{author}{Chen, Y.T.},
  \bibinfo{author}{Yeh, J.H.}, \bibinfo{author}{Li, P.J.},
  \bibinfo{year}{2006}.
\newblock \bibinfo{title}{Mandarin emotional speech recognition based on {SVM}
  and {NN}}, in: \bibinfo{booktitle}{18th International Conference on Pattern
  Recognition (ICPR'06)}, pp. \bibinfo{pages}{1096--1100}.
\bibitem[{Rabiner and Sambur(1975)}]{6778857}
\bibinfo{author}{Rabiner, L.R.}, \bibinfo{author}{Sambur, M.R.},
  \bibinfo{year}{1975}.
\newblock \bibinfo{title}{An algorithm for determining the endpoints of
  isolated utterances}.
\newblock \bibinfo{journal}{The Bell System Technical Journal}
  \bibinfo{volume}{54}, \bibinfo{pages}{297--315}.
\bibitem[{Rusu et~al.(2016)Rusu, Rabinowitz, Desjardins, Soyer, Kirkpatrick,
  Kavukcuoglu, Pascanu and Hadsell}]{DBLP:journals/corr/RusuRDSKKPH16}
\bibinfo{author}{Rusu, A.A.}, \bibinfo{author}{Rabinowitz, N.C.},
  \bibinfo{author}{Desjardins, G.}, \bibinfo{author}{Soyer, H.},
  \bibinfo{author}{Kirkpatrick, J.}, \bibinfo{author}{Kavukcuoglu, K.},
  \bibinfo{author}{Pascanu, R.}, \bibinfo{author}{Hadsell, R.},
  \bibinfo{year}{2016}.
\newblock \bibinfo{title}{Progressive neural networks}.
\newblock \bibinfo{journal}{CoRR} \bibinfo{volume}{abs/1606.04671}.
\bibitem[{Sahu et~al.(2017)Sahu, Gupta, Sivaraman, Espy-Wilson and
  AbdAlmageed}]{10995}
\bibinfo{author}{Sahu, S.}, \bibinfo{author}{Gupta, R.},
  \bibinfo{author}{Sivaraman, G.}, \bibinfo{author}{Espy-Wilson, C.},
  \bibinfo{author}{AbdAlmageed, W.}, \bibinfo{year}{2017}.
\newblock \bibinfo{title}{Adversarial auto-encoders for speech based emotion
  recognition}, in: \bibinfo{booktitle}{InterSpeech 2017}, pp.
  \bibinfo{pages}{1243--1247}.
\bibitem[{Shen et~al.(2011)Shen, Changjun and Chen}]{6023178}
\bibinfo{author}{Shen, P.}, \bibinfo{author}{Changjun, Z.},
  \bibinfo{author}{Chen, X.}, \bibinfo{year}{2011}.
\newblock \bibinfo{title}{Automatic speech emotion recognition using support
  vector machine}, in: \bibinfo{booktitle}{Proceedings of 2011 International
  Conference on Electronic Mechanical Engineering and Information Technology},
  pp. \bibinfo{pages}{621--625}.
\bibitem[{Sokolova and Lapalme(2009)}]{SOKOLOVA2009427}
\bibinfo{author}{Sokolova, M.}, \bibinfo{author}{Lapalme, G.},
  \bibinfo{year}{2009}.
\newblock \bibinfo{title}{A systematic analysis of performance measures for
  classification tasks}.
\newblock \bibinfo{journal}{Information Processing \& Management}
  \bibinfo{volume}{45}, \bibinfo{pages}{427 -- 437}.
\bibitem[{Ververidis et~al.(2004)Ververidis, Kotropoulos and Pitas}]{1326055}
\bibinfo{author}{Ververidis, D.}, \bibinfo{author}{Kotropoulos, C.},
  \bibinfo{author}{Pitas, I.}, \bibinfo{year}{2004}.
\newblock \bibinfo{title}{Automatic emotional speech classification}, in:
  \bibinfo{booktitle}{2004 IEEE International Conference on Acoustics, Speech,
  and Signal Processing}, pp. \bibinfo{pages}{I--593--6 vol.1}.
\bibitem[{Viola and Jones(2004)}]{Viola:2004:RRF:966432.966458}
\bibinfo{author}{Viola, P.}, \bibinfo{author}{Jones, M.J.},
  \bibinfo{year}{2004}.
\newblock \bibinfo{title}{Robust real-time face detection}.
\newblock \bibinfo{journal}{Int. J. Comput. Vision} \bibinfo{volume}{57},
  \bibinfo{pages}{137--154}.
\bibitem[{Xiao et~al.(2005)Xiao, Dellandrea, Dou and Chen}]{1577304}
\bibinfo{author}{Xiao, Z.}, \bibinfo{author}{Dellandrea, E.},
  \bibinfo{author}{Dou, W.}, \bibinfo{author}{Chen, L.}, \bibinfo{year}{2005}.
\newblock \bibinfo{title}{Features extraction and selection for emotional
  speech classification}, in: \bibinfo{booktitle}{IEEE Conference on Advanced
  Video and Signal Based Surveillance, 2005.}, pp. \bibinfo{pages}{411--416}.
\bibitem[{Zhang et~al.(2018)Zhang, Huang and Gao}]{8085174}
\bibinfo{author}{Zhang, S.}, \bibinfo{author}{Huang, T.}, \bibinfo{author}{Gao,
  W.}, \bibinfo{year}{2018}.
\newblock \bibinfo{title}{Speech emotion recognition using deep convolutional
  neural network and discriminant temporal pyramid matching}.
\newblock \bibinfo{journal}{IEEE Transactions on Multimedia}
  \bibinfo{volume}{20}, \bibinfo{pages}{1576--1590}.
\bibitem[{Zhang et~al.(2017)Zhang, Huang, Gao and Tian}]{7956190}
\bibinfo{author}{Zhang, S.}, \bibinfo{author}{Huang, T.}, \bibinfo{author}{Gao,
  W.}, \bibinfo{author}{Tian, Q.}, \bibinfo{year}{2017}.
\newblock \bibinfo{title}{Learning affective features with a hybrid deep model
  for audio-visual emotion recognition}.
\newblock \bibinfo{journal}{IEEE Transactions on Circuits and Systems for Video
  Technology} \bibinfo{volume}{PP}, \bibinfo{pages}{1--1}.

\end{thebibliography}

\end{document}